\documentclass[10pt,twocolumn,letterpaper]{article}

\usepackage[pagenumbers]{iccv} 

\usepackage{multirow}
\usepackage{multicol}
\usepackage{arydshln}
\usepackage{graphicx}
\usepackage{subcaption}
\usepackage{float}
\usepackage{enumitem}
\usepackage{url}
\usepackage{xcolor}
\usepackage{tikz}
\usepackage{setspace}
\usepackage{amsmath}
\usepackage{pifont}
\usepackage{textcomp}
\usepackage{booktabs}

\definecolor{iccvblue}{rgb}{0.21,0.49,0.74}
\usepackage[pagebackref,breaklinks,colorlinks,allcolors=iccvblue]{hyperref}

\def\paperID{11226} 
\def\confName{ICCV}
\def\confYear{2025}

\begin{document}

\title{Explainable Synthetic Image Detection through Diffusion Timestep Ensembling}


\author{
    \textbf{Yixin Wu} \thanks{Equal contribution.}\textbf{,} \
    \textbf{Feiran Zhang} \footnotemark[1]\textbf{,} \
    \textbf{Tianyuan Shi} \footnotemark[1]\textbf{,} \
    \textbf{Ruicheng Yin,} \
    \textbf{Zhenghua Wang,} \
    \textbf{Zhenliang Gan,} \\
    \textbf{Xiaohua Wang,} \
    \textbf{Changze Lv,} \
    \textbf{Xiaoqing Zheng} \thanks{Corresponding Author.}\textbf{,} \
    \textbf{Xuanjing Huang} \\
    School of Computer Science, Fudan University, Shanghai, China \\
    Shanghai Key Laboratory of Intelligent Information Processing \\
    {\tt\small $\{$yixinwu23$\}$m.fudan.edu.cn}
    {\tt\small $\{$zhengxq,xjhuang$\}$@fudan.edu.cn}
}

\maketitle
\begin{abstract}
Recent advances in diffusion models have enabled the creation of deceptively real images, posing significant security risks when misused. In this study, we empirically show that different timesteps of DDIM inversion reveal varying subtle distinctions between synthetic and real images that are extractable for detection, in the forms of such as Fourier power spectrum high-frequency discrepancies and inter-pixel variance distributions. Based on these observations, we propose a novel synthetic image detection method that directly utilizes features of intermediately noised images by training an ensemble on multiple noised timesteps, circumventing conventional reconstruction-based strategies. To enhance human comprehension, we introduce a metric-grounded explanation generation and refinement module to identify and explain AI-generated flaws. Additionally, we construct the GenHard and GenExplain benchmarks to provide detection samples of greater difficulty and high-quality rationales for fake images. Extensive experiments show that our method achieves state-of-the-art performance with 98.91\% and 95.89\% detection accuracy on regular and challenging samples respectively, and demonstrates generalizability and robustness. Our code and datasets are available at \href{https://github.com/Shadowlized/ESIDE}{https://github.com/Shadowlized/ESIDE}.
\end{abstract}
\section{Introduction}
\label{sec:introduction}

With the booming development of diffusion models such as Stable Diffusion \cite{Rombach2021HighResolutionIS}, DALL-E 3 \cite{betker2023improving}, Midjourney \cite{midjourney2022} and Flux, the proliferation of artificially generated images has reached unprecedented levels. While users marvel at the stunning quality of these synthetic visuals, a growing conundrum has also risen: distinguishing these creations from genuine photographs has become increasingly difficult, and the risks of malicious use have also skyrocketed. Can prevailing detection methods keep pace with the sophistication of these knockoffs? Moreover, can current detectors provide explanations robust enough to satisfy skeptics with more than just a feeble \textit{yes} or \textit{no}? Existing methods fail to cover more challenging images, and human comprehensibility of detection results is yet to be fully explored. We seek to address these gaps, by improving performance on harder detection samples and integrating high-quality explanations into our pipeline.

\begin{figure}[t!]
    \centering
    \includegraphics[width=\linewidth]{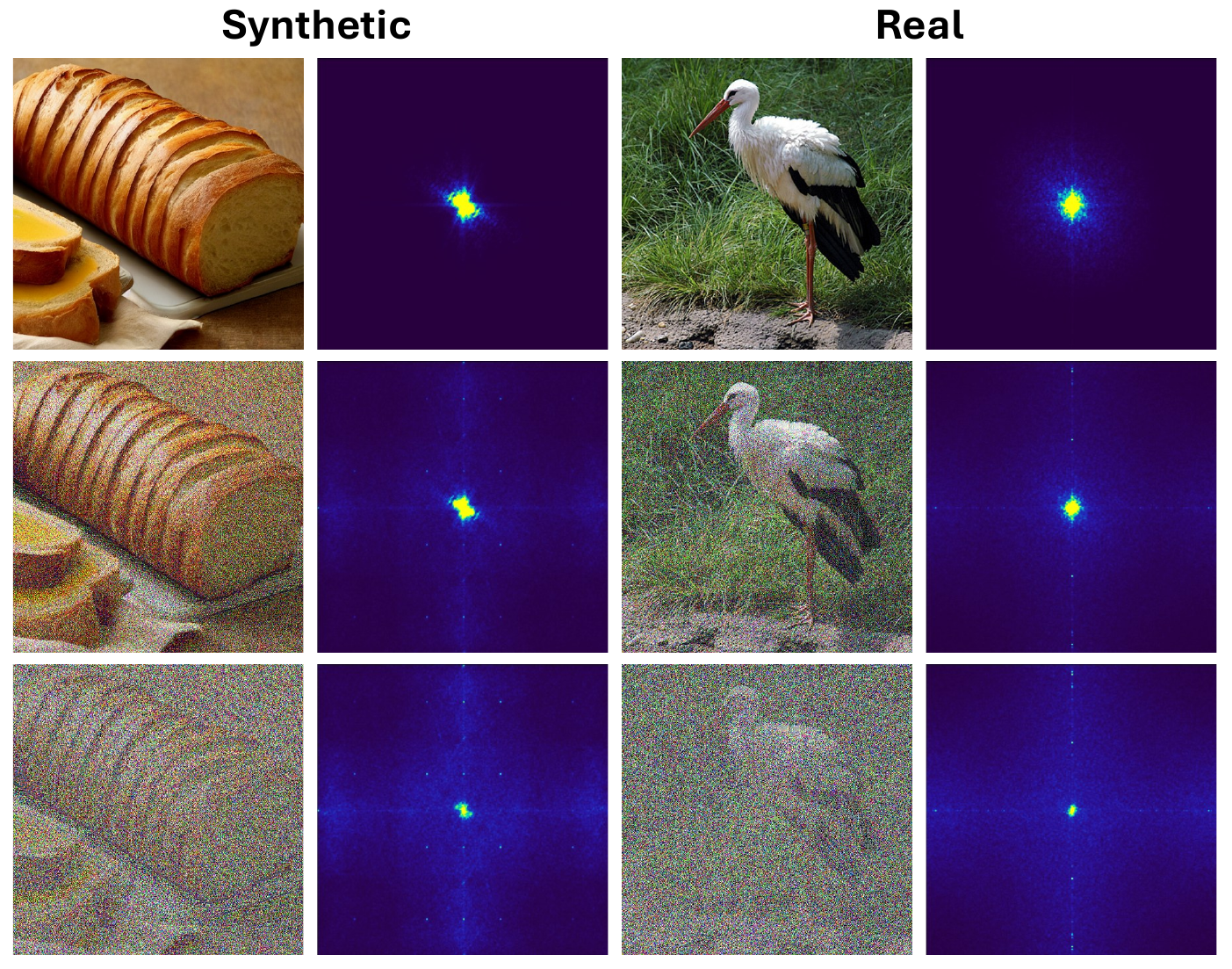}


    \caption{Fourier power spectra of synthetic and real images at timestep 0, 6, 12 of a 24-timestep DDIM inversion process. Artifacts of synthesized images manifest as peaks in the high-frequency components of the spectral background, becoming more pronounced with increasing timesteps.}
    \vspace{-3mm}

    \label{fig:fft_diffusion}
\end{figure}

\begin{figure*}[ht] 
    \centering
    \includegraphics[width=\textwidth]{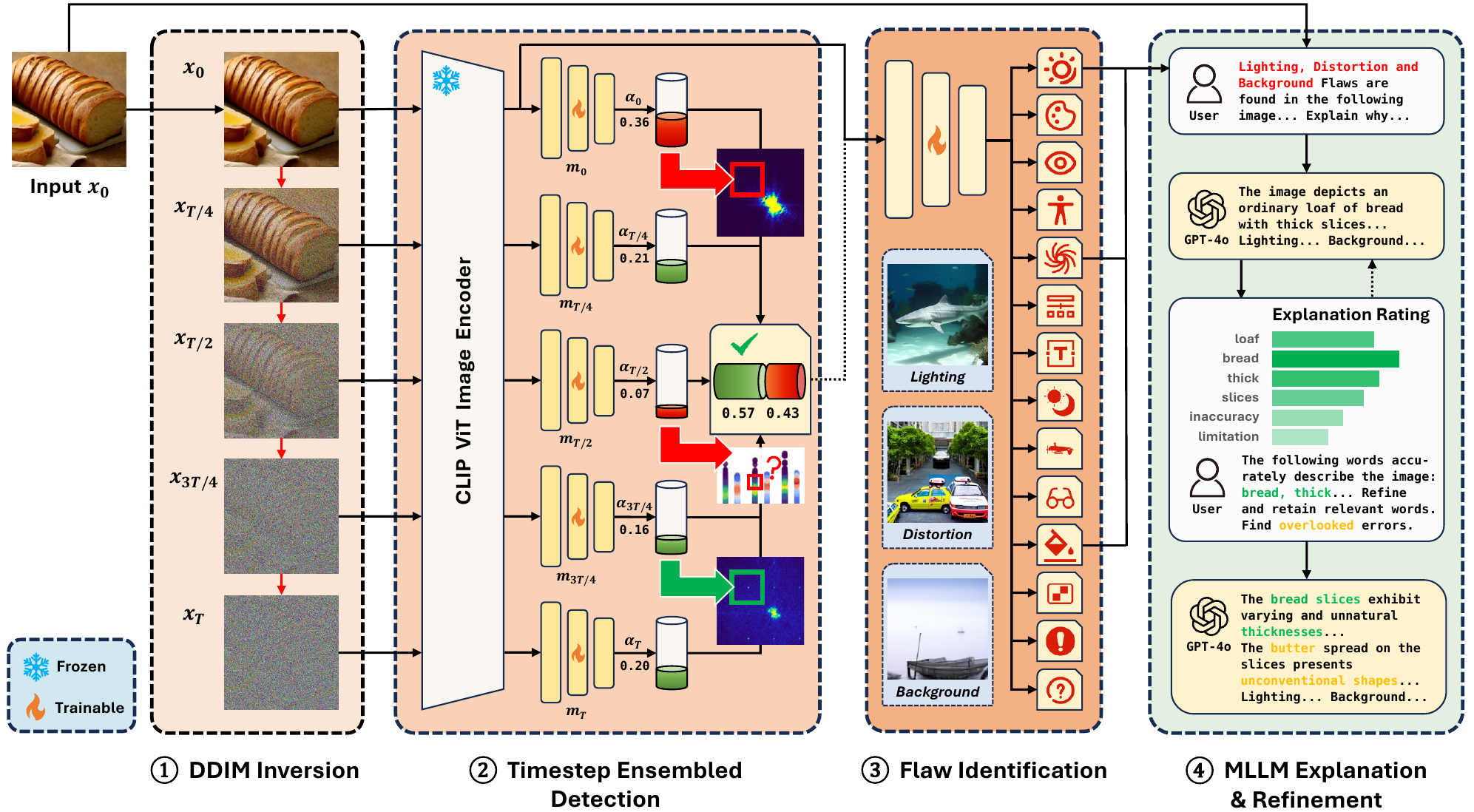}

    \caption{Illustration of the \textbf{ESIDE} pipeline. \textbf{Stage 1:} DDIM inversion progressively adds noise to input images, creating intermediately noised images. \textbf{Stage 2:} Synthetic image detection based on ensembling noised timesteps, discriminators are trained on noised images corresponding to distinct diffusion-induced data distributions, capturing various intermediate features. \textbf{Stage 3:} Multi-label flaw identification for synthetic images. \textbf{Stage 4:} Explanation generation with MLLMs and rated refinement.}

    \label{fig:main_figure}
\end{figure*}

Previous studies on synthetic image detection have employed deep neural networks \cite{Wang2019CNNGeneratedIA,tan2023learning,sha2023fake,cozzolino2024raising}, or exploited distinguishable fingerprints in the frequency and spatial domains \cite{dzanic2020fourier,liu2022detecting,corvi2023intriguing,zhong2023rich} to determine the fidelity of images.
Methods utilizing diffusion-based characteristics such as DIRE, SeDID, LaRE and DRCT \cite{Wang2023DIREFD,ma2023exposing,luo2024lare,chen2024drct} focus on detecting discrepancies by reconstructing images through noising and denoising processes, and identifying synthetic content via reconstruction errors.
These works inherently require \textit{\textbf{both}} forward and reverse processes, as their strategies rely on diffusion-based reconstruction.

\begin{figure}[b!] 
    \centering
    \includegraphics[width=\linewidth]{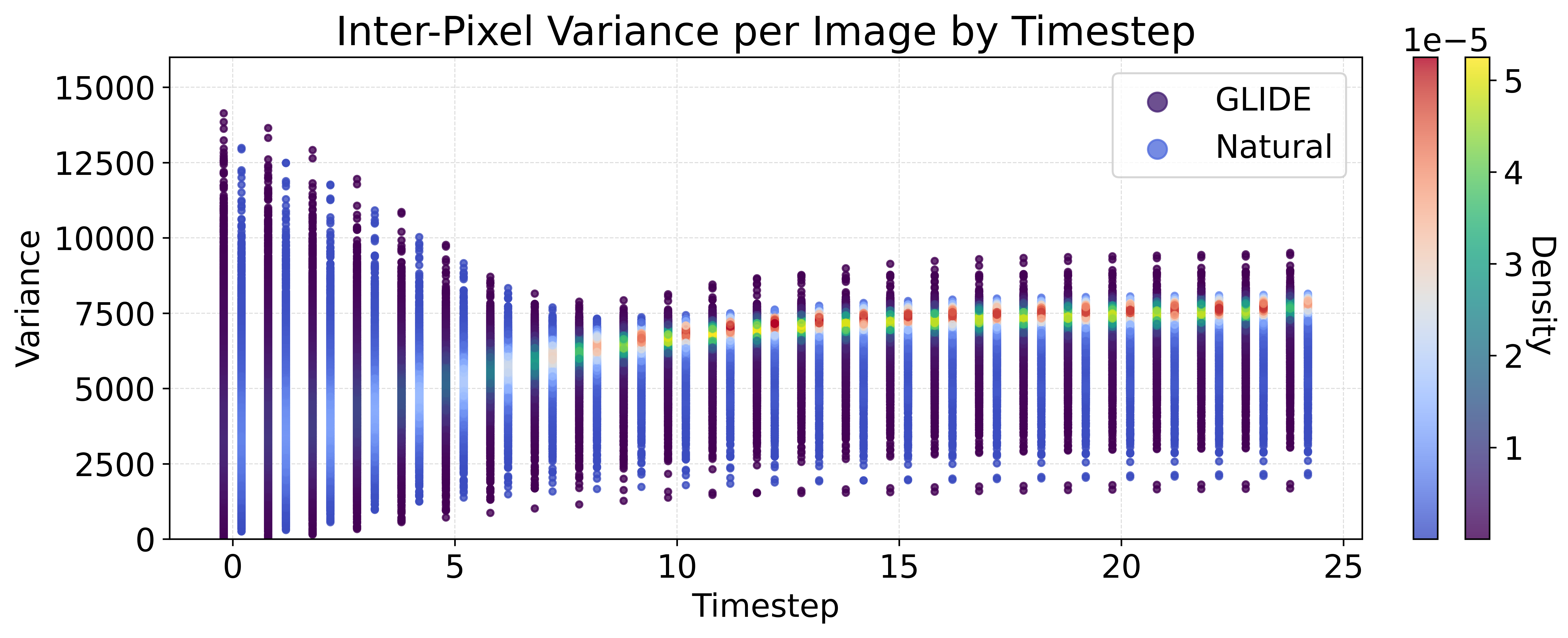}

    \caption{Inter-pixel variances of GLIDE-generated images and natural images noised through DDIM inversion.}

    \label{fig:variance_figure}
\end{figure}

However, features useful for detection in partially-noised images exist and are often overlooked.
As shown in Figure~\ref{fig:fft_diffusion}, synthetic and real images exhibit distinctions in high-frequency components of their Fourier spectra and manifest differing spectral structures throughout Gaussian diffusion-based noising. 
Since Fourier transform inherently captures transformation-invariant cues useful for image analysis \cite{reddy1996fft} and is mathematically connected to convolution via the convolution theorem, these frequency-domain features can be directly extracted.
Inspired by \cite{li2023alleviating}, we also observe that these images exhibit inconsistent inter-pixel variance distributions across timesteps in terms of spread and peak intensity, depicted in Figure~\ref{fig:variance_figure}. 
Synthetic and real images demonstrate disparate characteristics at each timestep, providing additional clues possibly usable for detection. 

Therefore, we bypass conventional reconstruction measures and propose a pipeline requiring only a \textit{\textbf{single}} noising pass, aiming to directly utilize subtle features within each intermediate noised step, named \textbf{ESIDE}: \textbf{E}xplainable \textbf{S}ynthetic \textbf{I}mage Detection through \textbf{D}iffusion Timestep \textbf{E}nsembling, as illustrated in Figure~\ref{fig:main_figure}. 
Our framework is designed for detecting generated images of greater challenge and synthesizing high-quality explanations. 
By circumventing conventional denoising, we halve the time consumed by DDIM \cite{songdenoising} pre-processing.
Previously underexplored research gaps regarding human comprehension of synthetic images are also addressed, grounding content generated by multimodal large language models (MLLMs) on computable evaluation metrics.

\vspace{1mm}
Our main contributions are four-fold:
\begin{itemize}[leftmargin=3.0mm, itemsep=0.3mm]
    \item We propose a performance-oriented synthetic image detection method based on ensemble learning of diffusion-noised images named ESIDE, achieving state-of-the-art performance on both regular and harder samples, while further pushing the limits of detectors to more challenging samples---those that truly demand reliable detection.

    \item We circumvent conventional image reconstruction measures and provide the insight that varying DDIM inversion intermediate timesteps reveal features directly extractable for detection. To the best of our knowledge, we are the first to systematically explore this approach.
    
    \item We introduce an explanation and refinement module into our pipeline for generating precise rationales, bridging the gap of underexplored explainability of generations.
    
    \item We construct two datasets: \scalebox{1.1}{G}\scalebox{0.8}{EN}\scalebox{1.1}{H}\scalebox{0.8}{ARD} and \scalebox{1.1}{G}\scalebox{0.8}{EN}\scalebox{1.1}{E}\scalebox{0.8}{XPLAIN}, providing researchers access to images of greater detection difficulty, with synthetic flaws and explanations.
\end{itemize}
\section{Related Work}
\label{sec:related_work}

\subsection{Synthetic Image Detection}

Methods that analyze standalone image characteristics without supplementary captions or additional contextual information for synthetic detection could be broadly categorized into three main approaches \cite{laurier2024cat}: deep learning detectors, frequency analysis and spatial analysis.

Deep learning-based detection methods are the most commonly adopted. 
Early detectors primarily targeted images generated by traditional convolutional neural networks \cite{Wang2019CNNGeneratedIA}.
More recently, methods leveraging vision transformers (ViTs) gained prominence, utilizing CLIP-ViT \cite{Radford2021LearningTV} image encoders for feature extraction, combined with additional classifiers networks or similarity metrics \cite{ojha2023towards,sha2023fake,cozzolino2024raising,lin2024robust,xu2024famsec}. 
LGrad used pretrained StyleGAN to convert images into gradients \cite{tan2023learning}.
Other studies that leverage diffusion methods \cite{Wang2023DIREFD,ma2023exposing,luo2024lare,chen2024drct} mainly focus on identifying synthetical discrepancies by comparing images with their reconstructions through diffusion noising and denoising.

On the other hand, frequency analysis generally classify synthetic images based on their high-frequency features. Generated images share systematic shortcomings in replicating attributes of high-frequency Fourier modes \cite{dzanic2020fourier,corvi2023intriguing}. Frequency inconsistencies and patterns among images generated with different models could also be effectively utilized for detection \cite{liu2024hfnet, song2024trinity, tan2024frequency}. 
However, utilizing the amplification of distinctions between synthetic and authentic images through intermediate steps of DDIM inversion is yet to be investigated.

Meanwhile, spatial analysis methods detect fake images by computing pixel-level relations and noise patterns. Inter-pixel relationships and contrasts could be captured and used to train detectors \cite{zhong2023rich,tan2024rethinking}, while noise patterns of real and synthetic images extracted through spatial models exhibit distinct characteristics usable for classification \cite{liu2022detecting,chen2024single}.

\vspace{1mm}
\subsection{Detection Explainability}

Previous studies have explored image forgery explanation using MLLMs \cite{huang2024ffaa,Xu2024FakeShieldEI,kang2025legion}, focusing on identifying manipulations and modifications rather than explaining images synthesized from scratch, lying outside our scope of discussion. 
Existing synthetic explainers introduce benchmarks rather limited in size, and rely entirely on MLLMs for detection without integrating specialized models and metrics \cite{li2024fakebench,ye2024loki}. Latest work \cite{wen2025spot} utilizes LLaVA \cite{liu2023visual} for detection, but requires substantial training resources for moderate performance, while the proposed benchmark broadly categorizes based on image content rather than the actual reason it is identified as synthetic.

Instead, our work proposes a light-weighted unified framework combining detection, explanation and automated refinement. We introduce an explanation benchmark for synthetic images named \scalebox{1.1}{G}\scalebox{0.8}{EN}\scalebox{1.1}{E}\scalebox{0.8}{XPLAIN}, larger in scale than current benchmarks and categorizing images by their actual synthetical appearance. Manual data pruning revealed a maximal of 67.4\% of identified flaws to be incorrect, demonstrating that exclusive reliance on MLLMs proves untrustworthy. By anchoring explanations in classified synthetical errors, and incorporating a refinement process guided by quantitative metrics, we effectively mitigate the limitations of MLLMs in standalone detection tasks.

\section{Method}
\label{sec:method}

\subsection{Preliminaries}
Diffusion models \cite{Rombach2021HighResolutionIS,betker2023improving,midjourney2022} generate images through a two-stage process of forward noise addition and reverse denoising. The forward process gradually transforms data into Gaussian noise, while the reverse trains a neural network to iteratively denoise for distribution restoration. Diffusion models build a symmetric Markov chain connecting the processes, aiming to minimize the KL divergence between data and noise distributions.

\vspace{1mm}

The forward process of the Markov chain is defined as:
\vspace{2mm}
\begin{equation}
    \small
q(x_t | x_{t-1}) = \mathcal{N}\left(x_t; \sqrt{1-\beta_t}x_{t-1}, \beta_t \mathbf{I}\right),
  \label{eq:DM}
\end{equation}
where $x_t$ represents the noised image at timestep $t$, and $\beta_t \in (0,1)$ is a predefined noise schedule.

\vspace{1mm}

\noindent\textbf{Denoising Diffusion Probabilistic Models (DDPM)} \cite{sohl2015deep} parameterize the reverse process as a Markov chain:
\begin{equation}
  \vspace{3mm}
    \small
p_\theta(x_{t-1} | x_t) = \mathcal{N}\left(x_{t-1}; \mu_\theta(x_t, t), \sigma_t^2 \mathbf{I}\right),
  \label{eq:DDPM1}
  \vspace{-2mm}
\end{equation}
where the mean \( \mu_\theta \) is derived from a neural network that predicts the noise \( \epsilon_\theta(x_t, t) \):  

\begin{equation}
  \vspace{3mm}
    \small
\mu_\theta(x_t, t) = \frac{1}{\sqrt{\alpha_t}} \left( x_t - \frac{\beta_t}{\sqrt{1-\bar{\alpha}_t}} \epsilon_\theta(x_t, t) \right).
  \label{eq:DDPM2}
  \vspace{1mm}
\end{equation}

\noindent\textbf{Denoising Diffusion Implicit Models (DDIM)} \cite{songdenoising} accelerate sampling by defining a non-Markovian forward process while maintaining the same marginal distribution \( q(x_t | x_0) \). The reverse process combines deterministic generation with stochastic noise injection:

\vspace{-4mm}
\begin{multline}
    \small
x_{t-1} = \sqrt{\bar{\alpha}_{t-1}} \left( \frac{x_t - \sqrt{1-\bar{\alpha}_t} \epsilon_\theta(x_t, t)}{\sqrt{\bar{\alpha}_t}} \right) +\\ \sqrt{1-\bar{\alpha}_{t-1} - \sigma_t^2} \cdot \epsilon_\theta(x_t, t) + \sigma_t \epsilon,
\label{eq:DDIM}
\end{multline}


\vspace{1mm}
\subsection{Synthetic Image Detection}

We propose a novel method for detecting synthetics through ensemble learning of intermediate noise. AdaBoost \cite{freund1997decision} is a boosting algorithm that aggregates the predictions of multiple weak models through a weighted sum, constructing a stronger learner with enhanced accuracy for discrimination. Specifically, we follow previous ensembling measures, but train classifiers on distinct data distributions, each corresponding to a different diffusion timestep.

Given an input image $x_0$, we apply a $T$-timestep DDIM inversion process to yield intermediate samples, generating a sequence of stepwise noised images: $\{x_0, x_1, x_2, \dots, x_T\}$.
For each timestep with an interval of a stride $s$, a base classifier $m_k$ is trained exclusively on corresponding timestep noised images, resulting in a collection of models $M$: 

\begin{equation}
  M = \{m_0, m_s, m_{2s}, \dots, m_T\}
  \label{eq:model_ensemble}
\end{equation}

A \textbf{sample weight} $w_{k,i}$ is assigned to each image of the training set to emphasize samples previously misclassified, while reducing the significance of correctly predicted cases. As each model operates on a noised dataset corresponding to a different diffusion timestep, the sample weights are separately initialized for each model, where $N$ represents the total number of images in the training set.

\begin{equation}
  w_{k,1} = w_{k,2} = \dots = w_{k,n} = \frac{1}{N}
  \label{eq:init_sample_weights}
\end{equation}

By incorporating $w_{k,i}$ into binary cross-entropy (BCE), a weighted loss function $\mathcal{L}_{\text{WBCE}}$ is obtained, where $y$ denotes the true label and $\hat{y}$ is the predicted probability:

\begin{equation}
    \small
    \hspace{-3mm}\mathcal{L}_{\text{WBCE}}(k, y, \hat{y}) = \sum_{i=1}^N w_{k,i} \left[ y_i \log(\hat{y}_i) + (1 - y_i) \log(1 - \hat{y}_i) \right]\hspace{-2mm}
  \label{eq:weighted_bce_loss}
\end{equation}

The weighted error $\epsilon_k$ of $m_k$ can then be calculated as follows, where $h_k(x_i)$ is the prediction of $m_k$ for $x_i$:

\begin{equation}
  \epsilon_k = \frac{\sum_{i=1,h_k(x_i) \neq y_i}^N w_{k,i}}{\sum_{i=1}^N w_{k,i}}
  \label{eq:calc_model_error}
\end{equation}

To form the final prediction, a \textbf{model weight} $\alpha_k$ is assigned to each classifier and updated throughout training:

\begin{equation}
  \alpha_k = \frac{1}{2} \ln(\frac{1 - \epsilon_k}{\epsilon_k})
  \label{eq:calc_model_weights}
\end{equation}

Sample weights are then adjusted according to the prediction results and normalized across timestep samples:

\begin{equation}
  \tilde{w}_{k,i} = w_{k,i} \cdot e^{-\eta \alpha_k \cdot h_k(x_i) y_i}
  \label{eq:update_model_weights}
\end{equation}

\begin{equation}
  w^{'}_{k,i} = \frac{\tilde{w}_{k,i}}{\sum_{i=1}^N \tilde{w}_{k,i}}
  \label{eq:normalize_model_weights}
\end{equation}

Here, $\eta$ is a learning rate factor applied to limit change rate.
After training the base classifiers, the model weights $\alpha_k$ are used to calculate a weighted sum $H(x_i)$ of the predictions from each $m_k$ for the final prediction of image $x_i$:

\begin{equation}
  H(x_i) = \text{sign}(\sum_{k=0}^{T} \alpha_{k} \cdot h_k(x_i))
  \label{eq:weighted_score_ensemble}
\end{equation}

In practice, a threshold is applied to the weighted error to normalize model weights and prevent degradation.

\begin{figure*}[ht] 
    \centering
    \includegraphics[width=\textwidth]{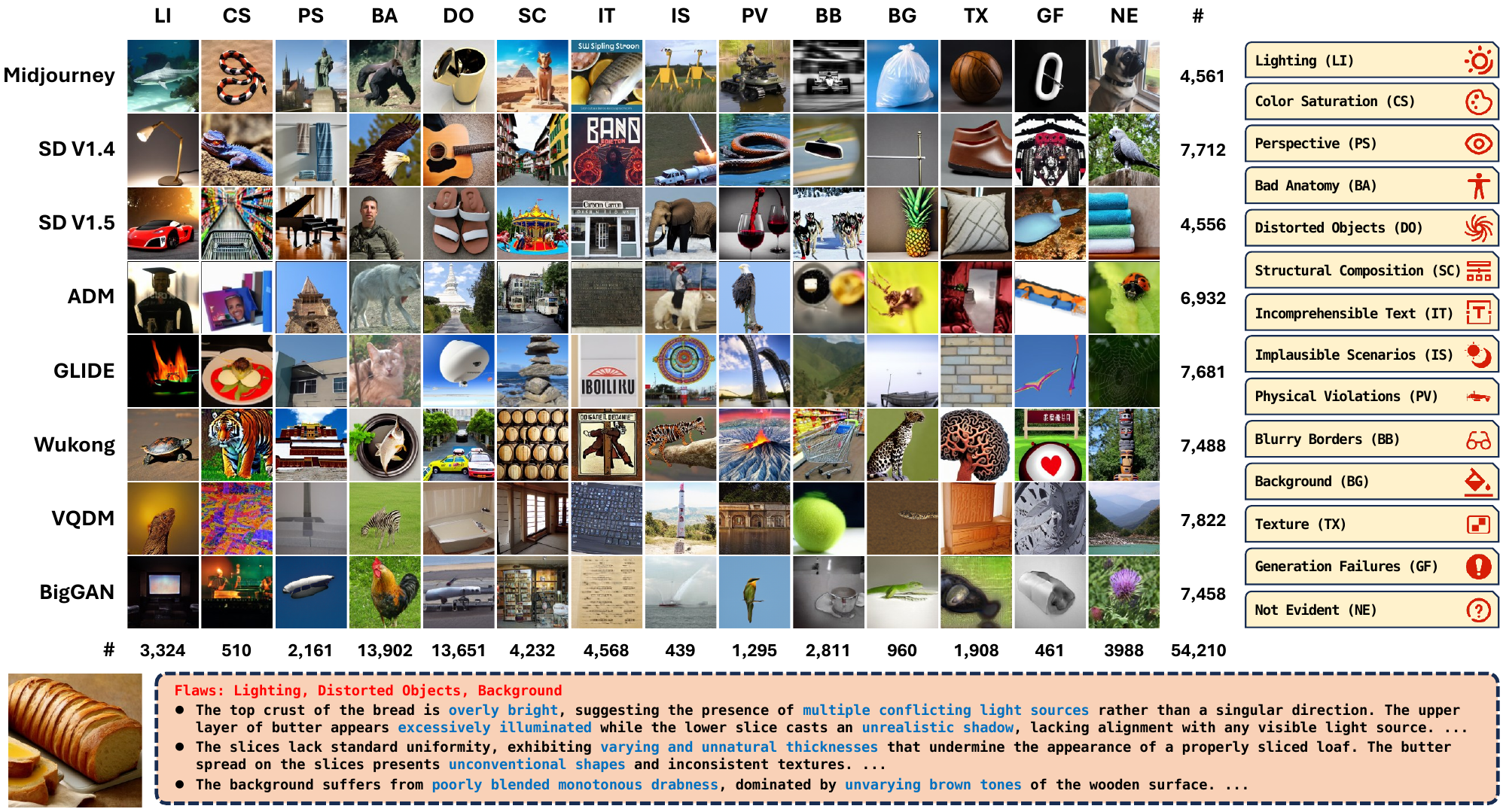}

    \caption{Visualization of the \textbf{GenExplain} benchmark. Synthetic images from GenImage are divided into 14 categories of flaws. Each image is matched with one or multiple categories, with a corresponding explanation for each flaw type.}

    \vspace{-3mm}
    \label{fig:explain_ds_figure}
\end{figure*}


\vspace{1mm}
\subsection{Multimodal Explanation Refinement}
\label{subsec:expl}
When an image is identified as synthetic, a multi-label classifier is then employed to identify potential flaws present. MLLMs are utilized to generate a rough explanation for each image based on the identified types. An refinement process is iteratively conducted to enhance explanation quality, segmenting the initial explanation into multiple phrases and assigning each a rating based on its semantic similarity with the image.
Text-image cross attention \cite{lee2018stacked} mechanisms are leveraged to compute these ratings. Faster R-CNN \cite{ren2015faster} is first applied for object detection, identifying $n$ sub-image regions, which are then encoded into visual embeddings $\{v_1, v_2, \dots, v_n\}$ using a CLIP ViT, while also concatenating the embedding of the full image $v_0$. Phrases $p$ are then encoded into the same vector space, and their similarities with each region are calculated and normalized, pairing these phrases with visual regions. The weighted combination of the region embeddings, denoted as $a$, is then derived based on the similarities:

\begin{equation}
    \tilde{s}_i=\frac{\cos\langle p,v_i\rangle}{\sum_{i=0}^n \cos\langle p,v_i\rangle}
    \label{eq:phrase_subregion_sim}
\end{equation}

\begin{equation}
    a=\sum^n_{i=0} {\frac{v_i \cdot e^{\lambda\tilde{s}_i}}{\sum_{i=0}^ne^{\lambda\tilde{s}_i}}}
    \label{eq:weighted_sum_subregion_sim}
\end{equation}
where $\lambda$ denotes the inverse temperature of the softmax function. The \textbf{rating} $r$ of the phrase $p$ is then calculated as its cosine similarity with $a$:

\begin{equation}
    r=\cos\langle p, a \rangle
    \label{eq:rating_calc_sim}
\end{equation}

For refinement, Top-K sampling is performed according to phrase relevance, and the MLLM is instructed to retain these phrases while identifying additional overlooked flawed regions. This process is iteratively repeated with the revised explanations, resulting in a final explanation that describes the flaws present in the image with greater accuracy.


\vspace{1mm}
\subsection{Construction of GenHard and GenExplain}

We construct two datasets, \scalebox{1.1}{G}\scalebox{0.8}{EN}\scalebox{1.1}{H}\scalebox{0.8}{ARD} and \scalebox{1.1}{G}\scalebox{0.8}{EN}\scalebox{1.1}{E}\scalebox{0.8}{XPLAIN}, based on GenImage \cite{zhu2023genimage}. The former comprises synthetic and natural images more challenging to detect, while the latter aims to categorize and provide explanations for flaws commonly found in artificial images.

\vspace{1mm}
\noindent \textbf{GenHard}\quad
To extract samples of greater difficulty, we employed CBSID \cite{cozzolino2024raising} with a minimalist linear network classifier under-fittingly trained for a single epoch on the validation subsets of GenImage, and subsequently tested on training subsets. Across the 8 subsets tested, the 108,704 synthetic images and 112,682 natural images misclassified were identified as hard samples, which were then partitioned into training and validation sets.

\vspace{1mm}
\noindent \textbf{GenExplain}\quad
Extending prior taxonomies \cite{mathys2024synthetic,li2024fakebench,Xu2024FakeShieldEI}, we identified 14 common categories of flaws associated with realistic synthetic images, and constructed a dataset comprising 54,210 groups of images, flaws and explanations, illustrated in Figure~\ref{fig:explain_ds_figure}. Images from GenImage validation subsets were fed into \texttt{gpt-4o}, prompted with flaw definitions and instructions, yielding a preliminary categorization of approximately 11,000 to 14,000 image-flaw pairs per subset. Manual data pruning removed 30.1\% to 67.4\% of images incorrectly categorized from each subset, and the final explanations are obtained through iterative refinement.

\begin{table*}[ht]
    \centering
    \small
    \setlength{\tabcolsep}{2.0pt}
    \renewcommand{\arraystretch}{1.2}
\resizebox{\textwidth}{!}{
    \begin{tabular}{p{1.4cm}ccccccccc}
    \toprule
\multirow{2}{*}{\textbf{Method}} & \multicolumn{8}{c}{\textbf{Training \& Test Subsets}} & \multirow{2}{*}{\textbf{Avg. Acc (\%)}} \\ \cline{2-9}
                        & \textbf{Midjourney} & \textbf{SD V1.4} & \textbf{SD V1.5} & \textbf{ADM} & \textbf{GLIDE} & \textbf{Wukong} & \textbf{VQDM} & \textbf{BigGAN} & \\ \midrule
ResNet-50 
& 83.90/86.75 & 80.63/94.17 & 76.17/90.19 & 75.81/91.00 & \textbf{95.82}/99.00 & 73.72/89.50 & 92.94/98.67 & \textbf{99.10}/96.08 & 84.76/93.17
 \\
DeiT-S 
& 58.29/76.08 & 73.92/81.08 & 74.32/78.75 & 63.37/68.42 & 72.84/93.83 & 71.01/83.83 & 66.87/77.50 & 46.23/76.00 & 65.86/79.44 \\
Swin-T 
& 70.81/91.00 & 74.63/88.08 & 78.02/90.00 & 70.30/79.50 & 85.19/97.83 & 74.92/85.92 & 86.96/90.17 & 88.24/\underline{99.67} & 78.63/90.27 \\
CNNSpot 
& 70.91/83.92 & 78.02/89.25 & 80.15/88.06 & 68.32/72.58 & 78.57/87.08 & 78.95/87.83 & 89.97/97.42 & 70.44/96.08 & 76.92/87.78 \\
CBSID 
& 67.10/93.25 & 73.03/90.92 & 72.22/93.63 & 94.20/\underline{99.83} & 88.53/\underline{99.17} & 75.67/92.75 & 84.10/98.17 & \underline{98.27}/\textbf{99.91} & 81.64/95.95 \\
DIRE 
& 88.86/92.83 & 95.72/97.42 & 96.02/96.25 & 90.52/94.67 & 81.40/\textbf{99.83} & 84.17/92.67 & 94.90/97.83 & 93.29/\underline{99.67} & 90.91/96.40 \\
LGrad 
& 72.25/87.33 & 72.28/83.92 & 76.45/84.37 & 70.86/79.17 & 82.19/96.00 & 65.50/79.75 & 74.80/81.08 & 83.74/94.58 & 74.76/85.78 \\
UnivFD 
& 40.80/87.75 & 41.92/89.33 & 35.40/88.25 & 65.28/85.58 & 79.13/94.25 & 65.35/89.42 & 67.84/91.92 & 71.07/94.75 & 58.35/90.16 \\
FreqNet 
& 87.13/94.33 & 93.47/94.58 & \textbf{96.73}/95.12 & 95.69/91.50 & 82.13/98.25 & 90.88/90.17 & \textbf{98.60}/95.42 & 85.12/99.17 & 91.22/94.82 \\
NPR 
& 85.10/87.25 & 89.90/95.25 & 95.23/97.12 & \underline{99.36}/99.75 & 86.29/92.08 & \textbf{96.94}/\textbf{98.42} & 93.74/97.33 & 92.87/99.58 & 92.43/95.85 \\
DRCT 
& \textbf{92.89}/\underline{97.50} & \textbf{97.51}/\textbf{100.00} & 96.10/\textbf{99.38} & 88.03/96.67 & 85.59/98.33 & 91.29/95.83 & 96.17/\textbf{100.00} & 97.93/99.17 & \underline{93.19}/\underline{98.36} \\
\hdashline
\textbf{ESIDE}
& \underline{92.38}/\textbf{98.42} & \underline{96.65}/\underline{99.17} & \textbf{96.73}/\underline{98.63} & \textbf{99.43}/\textbf{100.00} & \underline{94.98}/99.00 & \underline{91.50}/\underline{97.25} & \underline{97.90}/\underline{99.33} & 97.58/99.50 & \textbf{95.89}/\textbf{98.91} \\ \bottomrule
    \end{tabular}
}

\caption{
     Synthetic image detection accuracy on GenImage subsets. Models are trained individually on each GenImage subset, with the \textbf{original} samples training set only, while tested on both the \textbf{original} samples test set, and a previously unseen \textbf{hard} samples test set from GenHard. The prior number for each cell marks the test accuracy on the \textbf{hard} samples, while the posterior marks the test accuracy on the \textbf{original} samples. The best scores are highlighted in \textbf{bold}, and the second best are \underline{underlined}. 
}

\vspace{-3mm}
\label{tab:main_table}
\end{table*}

\section{Experiments}
\label{sec:experiments}


\subsection{Synthetic Image Detection}

We train an ensemble on groups of noised images derived from DDIM inversion intermediate timesteps, following the implementation provided by \cite{dhariwal2021diffusion,Wang2023DIREFD} and using pre-trained diffusion models of sizes $256\times256$ and $512\times512$. \texttt{CLIP ViT-L/14} \cite{Radford2021LearningTV} is employed to extract image features, which are passed through multi-layer perceptrons for classification. 
After each component model generates a prediction, the final prediction is obtained by computing a weighted sum based on their model weights $\alpha_k$.

We evaluate our results on GenImage \cite{zhu2023genimage}, a million-scale dataset covering 8 generator subsets: Midjourney \cite{midjourney2022}, Stable Diffusion V1.4 \cite{Rombach2021HighResolutionIS}, Stable Diffusion V1.5 \cite{Rombach2021HighResolutionIS}, ADM \cite{dhariwal2021diffusion}, GLIDE \cite{Nichol2021GLIDETP}, Wukong \cite{wukong2022}, VQDM \cite{Gu2021VectorQD}, and BigGAN \cite{brocklarge}. Due to computational resource constraints, we partition the validation subsets of GenImage by a 9:1 ratio for training and evaluation.

The baselines ResNet-50 \cite{He2015DeepRL}, DeiT-S \cite{Touvron2020TrainingDI}, Swin-T \cite{Liu2021SwinTH}, CNNSpot \cite{Wang2019CNNGeneratedIA}, CBSID \cite{cozzolino2024raising}, DIRE \cite{Wang2023DIREFD}, LGrad \cite{tan2023learning}, UnivFD \cite{ojha2023towards}, FreqNet \cite{tan2024frequency}, NPR \cite{tan2024rethinking}, and DRCT \cite{chen2024drct} (\textit{ICML 2024 Spotlight}) are compared, on both the original GenImage dataset and the more challenging samples curated in \scalebox{1.1}{G}\scalebox{0.8}{EN}\scalebox{1.1}{H}\scalebox{0.8}{ARD}. Two distinct scenarios are investigated: (1) train-test subsets from the same generator, and (2) train-test subsets sourced from different generators, enabling a comprehensive assessment of performance and generalizability.

\begin{table*}[ht]
    \centering
    \small
    \setlength{\tabcolsep}{2.0pt}
    \renewcommand{\arraystretch}{1.2}
\resizebox{\textwidth}{!}{
    \begin{tabular}{p{1.4cm}ccccccccc}
    \toprule
\multirow{2}{*}{\textbf{Method}} & \multicolumn{8}{c}{\textbf{Test Subsets}} & \multirow{2}{*}{\textbf{Avg. Acc (\%)}} \\ \cline{2-9}
                        & \textbf{Midjourney} & \textbf{SD V1.4} & \textbf{SD V1.5} & \textbf{ADM} & \textbf{GLIDE} & \textbf{Wukong} & \textbf{VQDM} & \textbf{BigGAN} & \\ \midrule
ResNet-50 
& 45.27/71.17 & 80.63/94.17 & 74.06/90.33 & \textbf{75.36}/48.17 & 24.67/57.25 & 76.91/\underline{90.33} & 49.76/\underline{59.75} & 26.58/41.75 & 56.66/69.12 \\
DeiT-S 
& 47.88/49.33 & 73.92/81.08 & 75.21/79.50 & 65.21/48.17 & 25.67/48.92 & 74.55/80.75 & 47.89/46.08 & 31.70/43.58 & 55.25/59.68 \\
Swin-T 
& 47.14/56.58 & 74.63/88.08 & 76.13/86.94 & 57.71/49.83 & 21.45/50.33 & 76.54/84.08 & 43.24/49.25 & 26.92/45.08 & 52.97/63.77 \\
CNNSpot 
& 49.28/56.58 & 78.02/89.25 & 78.34/86.19 & 72.77/48.50 & 26.36/51.50 & 74.77/83.42 & 49.39/50.25 & 27.13/47.83 & 57.01/64.19 \\
CBSID 
& 51.16/74.33 & 73.03/90.92 & 74.07/91.69 & 73.69/54.42 & 49.36/\textbf{74.83} & 73.41/78.33 & 52.32/\textbf{65.58} & 33.36/\underline{59.08} & 60.05/\underline{73.65} \\
DIRE 
& 67.38/62.08 & 95.72/97.42 & 95.59/\underline{96.75} & 49.01/30.50 & 31.38/17.17 & 62.18/56.50 & 33.80/29.25 & 37.58/19.50 & 59.08/51.15 \\
LGrad 
& 50.28/56.17 & 72.28/83.92 & 73.57/81.92 & \underline{74.19}/44.50 & 25.05/49.50 & 55.56/53.83 & 53.21/52.92 & 33.46/45.33 & 54.70/58.51 \\
UnivFD 
& 26.75/\textbf{82.42} & 41.92/89.33 & 38.66/79.17 & 61.83/47.08 & 25.12/71.58 & 58.17/72.17 & 54.36/51.58 & 29.75/49.58 & 42.07/67.87 \\
FreqNet 
& \textbf{68.96}/70.50 & 93.47/94.58 & 88.52/91.58 & 67.51/\textbf{63.92} & 48.81/\underline{72.58} & \underline{87.23}/78.83 & 53.61/56.25 & \underline{70.44}/\textbf{71.83} & \underline{72.32}/\textbf{75.01} \\
NPR 
& 47.35/53.75 & 89.90/95.25 & 96.01/93.50 & 70.95/\underline{55.42} & \underline{50.80}/65.58 & \textbf{95.53}/\textbf{96.08} & 58.27/49.33 & 41.61/46.08 & 68.80/69.37 \\
DRCT
& 51.52/46.17 & \textbf{97.51}/\textbf{100.00} & \textbf{97.51}/86.50 & 69.36/53.50 & \textbf{55.68}/67.42 & 86.42/89.50 & \textbf{62.91}/56.25 & 44.39/52.25 & 70.66/68.95 \\
\hdashline
\textbf{ESIDE}
& \underline{67.40}/\underline{74.33} & \underline{96.65}/\underline{99.17} & \underline{97.15}/\textbf{98.75} & 63.79/37.08 & 50.21/39.00 & 82.60/67.33 & \underline{59.73}/50.42 & \textbf{90.10}/53.67 & \textbf{75.95}/64.97 \\
\bottomrule
    \end{tabular}
}

\caption{
     Cross-validation accuracy on GenImage subsets. Models are trained on GenImage/SD V1.4 with the \textbf{original} samples training set only, while tested on both the \textbf{original} and \textbf{hard} samples test set from another generator subset from GenHard.
}

\vspace{-3mm}
\label{tab:cross_validation_table}
\end{table*}

\begin{table}[t!]
    \centering
    \small
    \setlength{\tabcolsep}{1.0pt}
    \renewcommand{\arraystretch}{1.1}
    
    \begin{tabular}{p{0.9cm}ccccccccc}
    \toprule 
\textbf{Metric} & \textbf{MJ} & \textbf{SD1.4} & \textbf{SD1.5} & \textbf{ADM} & \textbf{GLI} & \textbf{WK} & \textbf{VQ} & \textbf{BG} & \textbf{Avg.} \\ \midrule
   
EM   & 41.60 & 52.03 & 58.08 & 50.41 & 49.72 & 33.81 & 54.61 & 49.07 & 48.67 \\ 
mAP  & 34.70 & 35.71 & 33.01 & 40.04 & 34.57 & 33.80 & 54.05 & 34.50 & 37.55 \\ \bottomrule
    \end{tabular}

\caption{
     Performance of 14-type flaw classification on GenExplain subsets. Subset names are abbreviated.
}

\label{tab:flaw_cls_table}
\end{table}

\vspace{2mm}
\noindent \textbf{Implementation Details}\quad
A total of $T=24$ DDIM inversion timesteps are taken. The sample weights learning rate $\eta$ is set to $0.25$, and classifier error thresholds $\epsilon_k =\min(\max(\tilde{\epsilon_k}, 0.001), 0.5)$ are enforced to prevent model weight overflow and ensemble degradation. 
We select a stride $s=3$ resulting in an ensemble of $9$ classifiers, and only require a simple five-layer MLP classifier with an input dimension of 768, hidden layer dimensions of [1024, 512, 256, 128], and a scalar output for such performance, easily adaptable to larger networks. A batch normalization layer, a LeakyReLU activation with a negative slope of $0.1$, and a Dropout layer with a dropout rate of $0.5$ are sequentially applied after each linear layer. The AdamW optimizer with a learning rate of $1\times10^{-4}$ and weight decay of $5\times10^{-4}$ is employed, alongside our modified weighted binary cross-entropy loss function $\mathcal{L}_{\text{WBCE}}$.

Our method is implemented based on PyTorch, exhibits low GPU memory consumption, and enables simple hybrid parallelism as models trained on different noised datasets could easily be allocated to various devices and subsequently combined. 
Computational support is provided by NVIDIA L20 48GB GPUs. 
Training could be further accelerated when image features are precomputed and stored, which sums up as the main overtime.

\label{exp_setting}

\vspace{2mm}
\noindent \textbf{Result Analysis}\quad
As shown in Table~\ref{tab:main_table}, ESIDE achieves SOTA performance on both harder and original images, with an average absolute accuracy increase of at least 2.70\% and 0.55\% respectively, performing 15.10\% better than baseline averages on harder samples and leading by 7.28\% on originals. Cross-validation results in Table~\ref{tab:cross_validation_table} show generalizability across images synthesized by other models. 
Notably, some methods perform worse than random guessing on hard samples as only incorrect identifications are included in \scalebox{1.1}{G}\scalebox{0.8}{EN}\scalebox{1.1}{H}\scalebox{0.8}{ARD}, further underscoring its difficulty. Similar cross-validation results reveal that effective indicators for one generator may perform opposite on another. Since we emphasize training on controversial images, the anomaly of higher cross-validation accuracies on samples of greater difficulty is observed, as original scenarios that should have been simple to classify could be under-trained in comparison.


\vspace{1mm}
\subsection{Error Explanation and Refinement}

\begin{figure}[t!]  
    \centering
    \includegraphics[width=\linewidth]{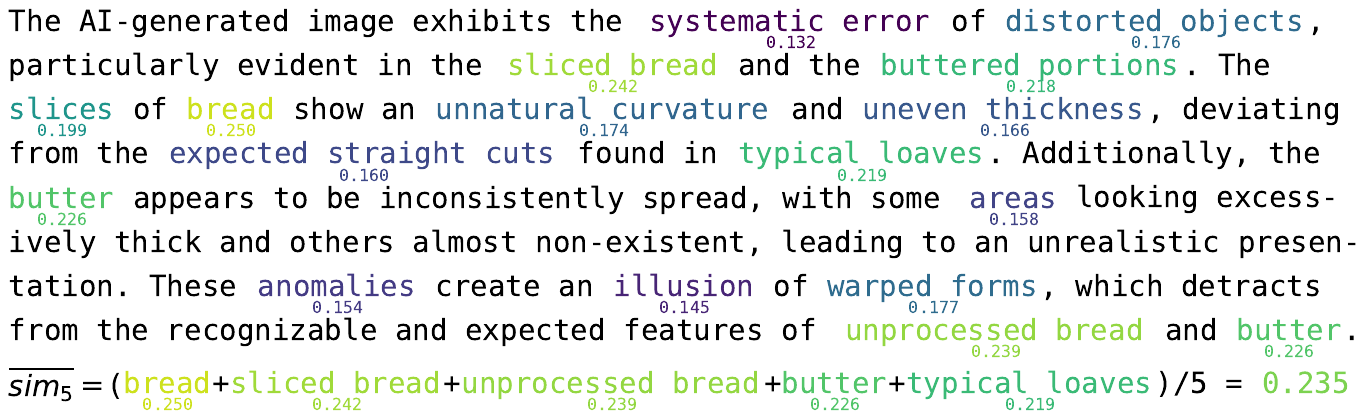}
    \caption{Illustration of phrase-image similarities in an explanation snippet and calculation of Top-5 average.}

    \vspace{-3mm}
    \label{fig:explanation_text}
\end{figure}

\noindent \textbf{Flaw Classification}\quad
We propose synthetic flaw detection as a new task, and introduce a simple baseline experimented on \scalebox{1.1}{G}\scalebox{0.8}{EN}\scalebox{1.1}{E}\scalebox{0.8}{XPLAIN}. The same classifier architecture as in Section~\ref{exp_setting} is adopted, with output layer dimension adjusted to 14 to match the categories of flaw types. BCE Loss is used, and each label is predicted independently. 
We partition our dataset by 9:1 for training and evaluation. 
Evaluation metrics include Exact Match (EM) accuracy and Mean Average Precision (mAP), measuring the proportion of predictions where all 14 labels are correctly matched, and the mean value of average precisions across all labels. 
Results are presented in Table~\ref{tab:flaw_cls_table}.

\begin{figure}[t!]  
    \centering
    \includegraphics[width=\linewidth]{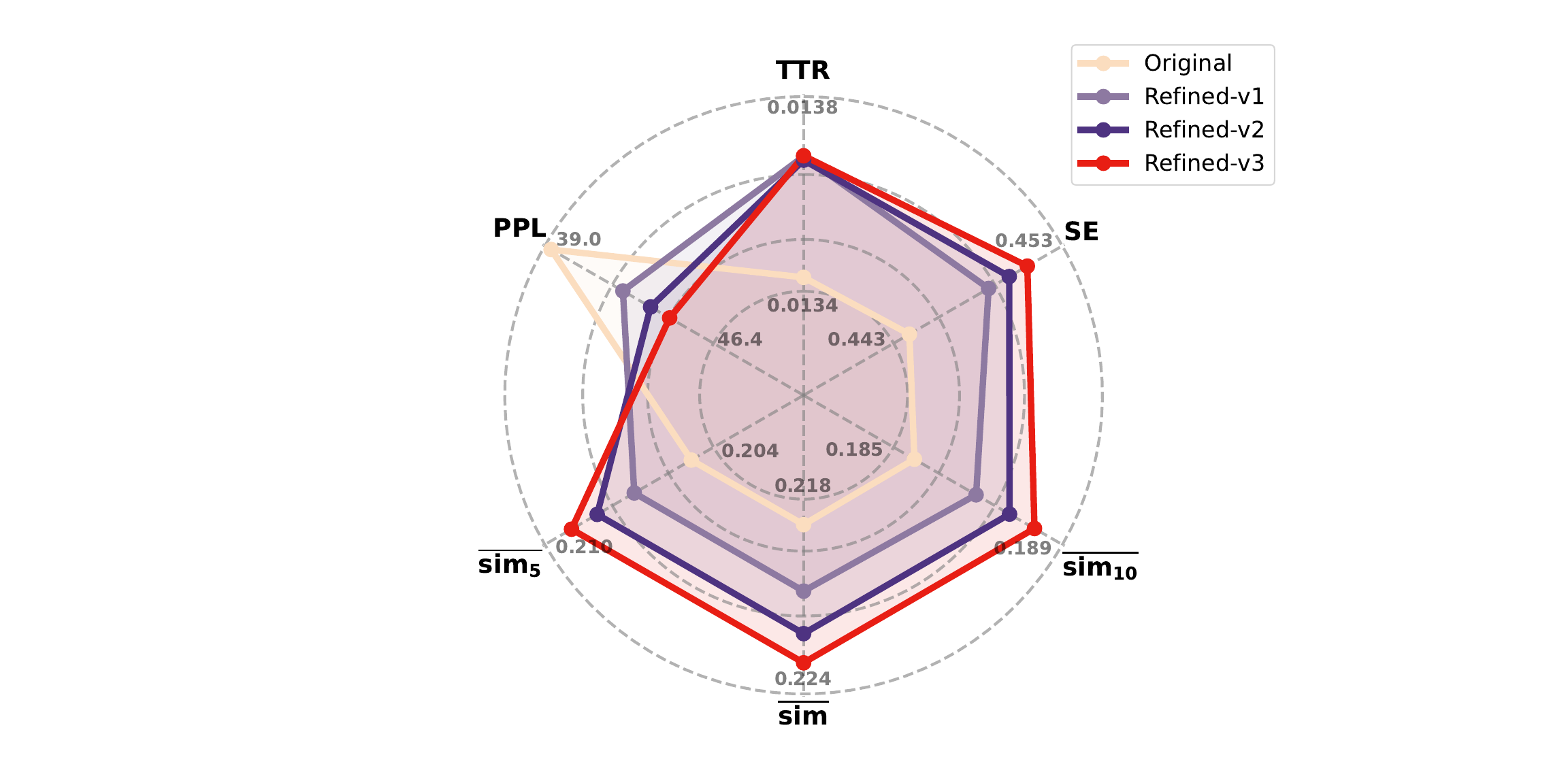}
    \caption{Metrics average across all 8 subsets of GenExplain regarding the initial explanations and their refined versions. }

    \vspace{-3mm}
    \label{fig:explanation_radar}
\end{figure}

\vspace{2mm}
\noindent \textbf{Explanation Refinement}\quad
We instruct \texttt{gpt-4o} to generate explanations, use \texttt{spacy} for phrase segmentation, and refine for 3 iterations. 
Top-5, Top-10, and Overall Similarity between text phrases and image regions are evaluated to guide refinement, and the top 10 phrases are retained, exemplified in Figure~\ref{fig:explanation_text}.
Additionally, Type-Token-Ratio (TTR$\uparrow$), normalized Shannon Entropy (SE$\uparrow$) and Perplexity (PPL$\downarrow$, tokenized using \texttt{gpt-2}) are employed to evaluate lexical diversity, information density and fluency, with results shown in Figure~\ref{fig:explanation_radar}.
Throughout refinement iterations, all similarity metrics, TTR and SE improved. However, PPL also rose due to the retention of specialized or domain-specific terms uncommon in general language usage.
Meanwhile, our \textit{Original} setting reflects the effect of baselines directly utilizing MLLMs for explanation, falling short on most metrics.

\vspace{1mm}
\subsection{Robustness Experiments}

In real-world scenarios, images awaiting detection often exhibit degradation. Following established procedures \cite{Wang2019CNNGeneratedIA,Wang2023DIREFD,lorenz2023detecting}, and additionally incorporating post-processing operations to better mimic actual cases, we assess the resilience of our method to distribution shifts by introducing three types of perturbations to test images: Gaussian blur ($\sigma\in[0.5, 2.5]$ pixels), arbitrary rotation ($\theta\in[-45^\circ, 45^\circ]$), and illumination variation (brightness factor $\alpha\in[0.3, 1.8]$). 
All models are trained on the original unmodified images for evaluation, and the best-performing baselines are compared.
As our method is exceptionally trained on more challenging samples, superior robustness is demonstrated, as observed in Table~\ref{tab:robustness_table}.

\begin{table}[t!]
    \centering
    \small
    \setlength{\tabcolsep}{1.0pt}
    \renewcommand{\arraystretch}{1.1}

    \begin{tabular}{p{1.1cm}ccccccccc}
    \toprule
\textbf{Method} & \textbf{MJ} & \textbf{SD1.4} & \textbf{SD1.5} & \textbf{ADM} & \textbf{GLI} & \textbf{WK} & \textbf{VQ} & \textbf{BG} & \textbf{Avg.}\\ \midrule
FreqNet & \textbf{85.67} & 84.50 & 84.94 & \textbf{70.25} & 77.50 & 74.25 & 73.25 & 76.08 & 78.31 \\
NPR     & 65.58 & 49.92 & 50.00 & 57.50 & 68.75 & 50.83 & 68.25 & \textbf{78.17} & 61.13 \\
DRCT    & \textbf{85.67} & 81.58 & \textbf{91.08} & 70.00 & 78.58 & 70.92 & 68.25 & 73.75 & 77.48 \\
\textbf{ESIDE} & 84.08 & \textbf{86.58} & 86.06 & 70.00 & \textbf{85.08} & \textbf{82.08} & \textbf{74.58} & 76.58 & \textbf{80.63} \\ \bottomrule
    \end{tabular}

\caption{
     Robustness performance evaluated on perturbated test subsets of GenImage. Subset names are abbreviated.
}

\label{tab:robustness_table}
\end{table}

\vspace{1mm}
\subsection{Ablation Studies}

\begin{table}[t!]
    \centering
    \small
    \setlength{\tabcolsep}{3.0pt}
    \renewcommand{\arraystretch}{1.1}

    \begin{tabular}{p{2.4cm}ccccc}
    \toprule
\textbf{Setting}     & \textbf{DDIM} & $\alpha_k$ & $w_{k,i}$ & \textbf{Acc (\%)}     & \textbf{$\Delta$}  \\ \midrule
\textbf{ESIDE}                & \ding{51}     & \ding{51}  & \ding{51} & \textbf{94.98}/99.00  & 0.00/0.00          \\
\hspace{0.3em}\textminus\hspace{0.1em} DDIM-Interm.  & \ding{51}     & \ding{51}  & \ding{51} & 94.37/98.00           & -0.61/-1.00         \\
\hspace{0.3em}\textminus\hspace{0.1em} DDIM       & \ding{55}     & \ding{51}  & \ding{51} & 89.27/99.00           & -5.71/0.00         \\
\hspace{0.3em}\textminus\hspace{0.1em} $\alpha_k$ & \ding{55}     & \ding{55}  & \ding{51} & 88.75/98.92           & -6.23/-0.08        \\
\hspace{0.3em}\textminus\hspace{0.1em} $\alpha_k,w_{k,i}$ & \ding{55}     & \ding{55}  & \ding{55} & 88.53/\textbf{99.17}  & -6.45/0.17         \\ \bottomrule
    \end{tabular}

\caption{
     Ablation studies of architectural components on detection results. Models are trained on GenImage/GLIDE. 
}

\label{tab:model_ablations_table}
\end{table}

\noindent \textbf{Noised Images and Ensembling}\quad
Would simply using unnoised or fully noised images perform better, and is performance increase merely due to the ensembling strategy?
To test this hypothesis, we evaluated four different settings correspondingly replacing partially-noised images of intermediate steps with fully-noised images, ensembling on unnoised images, removing ensembling, and eliminating misclassification-centric training. Table~\ref{tab:model_ablations_table} shows that our method achieves an accuracy increase of 5.71\% on hard samples compared to ensembling entirely on unnoised images, while deactivating model weights and sample weights further decreases performance. Ensembling a model trained on unnoised images with multiple models trained on fully noised images degrades performance on both distributions, implying that features from varying timesteps are utilized.

\vspace{2mm}
\noindent \textbf{High-Frequency Peaks Utilization}\quad
For both synthetic and natural images, we suppressed the highest percentile of their Fourier frequencies to a fixed ratio, while masking the commonly-shared components located within a specific bandwidth along the axes, and then used images reconstructed based on these modified spectra for training and evaluation. Table~\ref{tab:fft_modification_table} supports our insight that these frequency peaks could be captured by intermediate-step detectors to enhance detection capability regarding more questionable instances, as their suppression halves ensemble effect.

\begin{table}[t!]
    \centering
    \small
    \setlength{\tabcolsep}{3pt}
    \renewcommand{\arraystretch}{1.1}

    \begin{tabular}{ccccc}
    \toprule
\textbf{Bandwidth} & \textbf{Suppression} & \textbf{Percentile} & \textbf{Acc (\%)}     & \textbf{$\Delta$}  \\ \midrule
0.06     & 0.1  & 0.15 & 92.52/98.83 & -2.46/-0.17 \\
0.08     & 0.2  & 0.10 & 92.74/98.92 & -2.24/-0.08 \\ \bottomrule
    \end{tabular}

\caption{
    Effects of suppressing high-frequency peaks of Fourier power spectra quadrants. Models are trained and evaluated on reconstructions of GenImage/GLIDE.
}

\label{tab:fft_modification_table}
\end{table}
\vspace{1mm}
\section{Conclusion}
\label{sec:conclusion}

We present ESIDE, a novel pipeline for detecting and explaining synthetic images. We train an ensemble on noised images to directly utilize intermediate features introduced through DDIM inversion, circumventing conventional reconstruction measures. To improve human perception of fake images, we introduce an explanation generation and refinement module. Additionally, we construct two datasets, GenHard and GenExplain, comprising more challenging samples and providing categorized flaw types with explanations for AI-generated images. Extensive experiments show state-of-the-art performance on both regular and harder images, with significant improvements on tougher samples.
Our method also generalizes effectively, demonstrates robustness, and enables hybrid parallelism easily.

\vspace{1mm}
\section*{Limitations}
Our detection method circumvents conventional diffusion reconstruction, but still utilizes partially noised images that requires a rather time-consuming preceding DDIM inversion process.  
While our simple discriminator architecture enables faster training than existing detectors, the need to train multiple models for an ensemble cannot be denied, though being significantly less time-costing than DDIM inversion. 
Additionally, as we rely on MLLMs in our explanation and refinement process, explanation quality remains dependent on instruction-prompted MLLM performance. Throughout refining the \scalebox{1.1}{G}\scalebox{0.8}{EN}\scalebox{1.1}{E}\scalebox{0.8}{XPLAIN} dataset, 5 out of 6 quality metrics improved on average. However, it can not be guaranteed for each individual case that the MLLM performs effectively and generates higher-rated explanations, leading to rare loss of refinement effectiveness.

\vspace{1mm}
\section*{Societal Impacts}
Advances in diffusion models and AI-generated content have enabled the creation of deceptively real images, delighting users while simultaneously raising significant ethical concerns.
Our study offers a novel solution to the rapidly evolving landscape of synthetic image detection, aiming to address the societal challenges posed by the malicious use of AI-generated images and mitigate potential security risks associated with their proliferation.
We employ \texttt{gpt-4o} to generate explanations for synthetic images, which may occasionally produce uncontrollable generated content requiring further discrimination. Another point to note is that our \scalebox{1.1}{G}\scalebox{0.8}{EN}\scalebox{1.1}{H}\scalebox{0.8}{ARD} and \scalebox{1.1}{G}\scalebox{0.8}{EN}\scalebox{1.1}{E}\scalebox{0.8}{XPLAIN} datasets use images from the GenImage benchmark, which may include disturbing images with diffusion-generated malformed content, particularly in images regarding human faces or animals.

{
    \small
    \bibliographystyle{ieeenat_fullname}
    \bibliography{main}
}

\clearpage
\appendix

\newpage

\begin{figure*}[t!] 
    \centering
    \includegraphics[width=\textwidth]{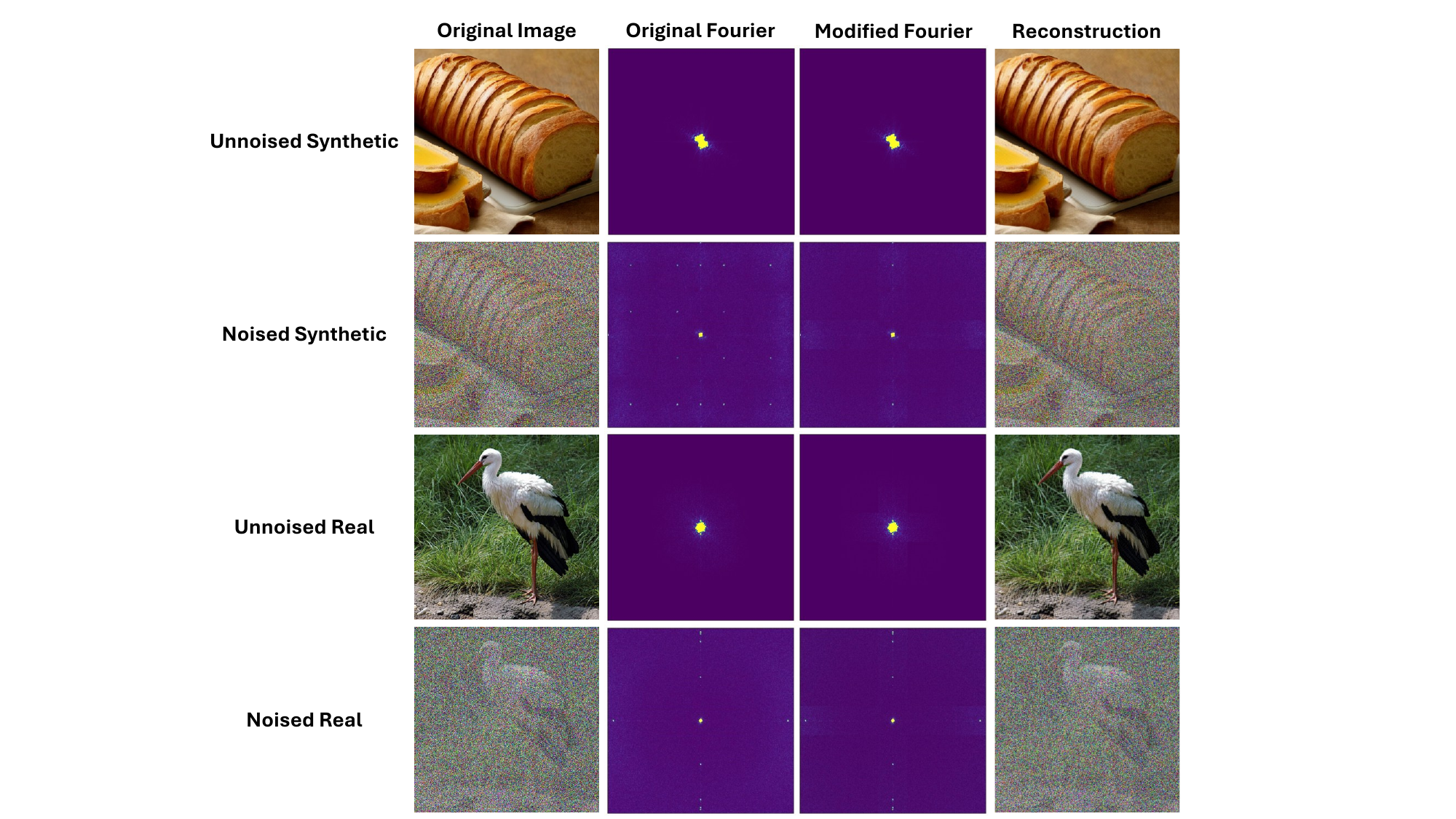}

    \caption{Fourier power spectra of synthetic and real images at timestep 0 and 12 of a 24-timestep DDIM inversion process. Energy levels of high-frequency peaks in the quadrants of the spectra are suppressed, and images are reconstructed based on these modified spectra and used for detection ablations. Synthetic images show prominent peaks in the quadrants, while real images show smooth and circular decrements.}

    \label{fig:fourier_recon_figure}
\end{figure*}

\begin{figure*}[t] 
    \centering
    \includegraphics[width=\textwidth]{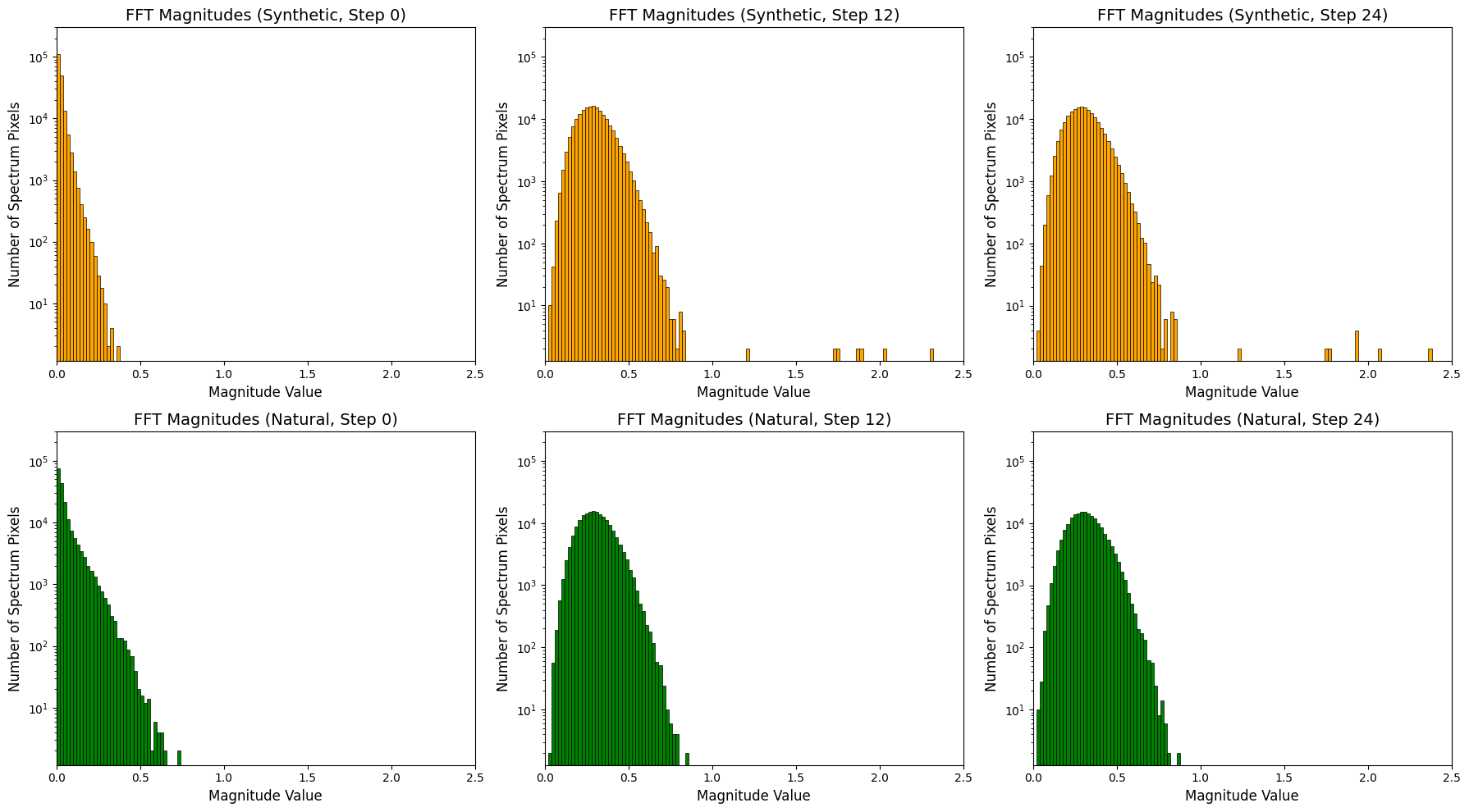}

    \caption{Energy magnitude histograms of Fourier power spectra of synthetic and real images at timestep 0, 12 and 24 of a 24-timestep DDIM inversion process. A bandwidth along the axes is masked. Synthetic images show high-frequency peaks gradually amplified by noising, unobserved in real images.}

    \label{fig:fourier_energy_histogram_figure}
\end{figure*}

\section{Experimental Details}\label{sec:a}
Further experimental settings, ablation studies, hyperparameter details, result analysis and questionable aspects are described and resolved in the sections below.

\subsection{Fourier Power Spectra Analysis}\label{sec:1.1}

For both synthetic and natural images, we suppress Fourier frequencies with the highest percentile of energy levels of the spectra to a small ratio of their original energy levels. We applied a mask of a specific bandwidth radius along the main horizontal and vertical axes by proportion of image width and height. Frequencies within this bandwidth are not recorded or suppressed, as both synthetic and natural images show high frequencies in these regions, which may obscure actual intriguing peaks. Synthetic and natural images were then reconstructed based on these modified spectra, and used for training and evaluation. An example of this process is depicted in Figure~\ref{fig:fourier_recon_figure}, while quantitative histograms of the energy levels of synthetic and natural images from varying diffusion timesteps are depicted in Figure~\ref{fig:fourier_energy_histogram_figure}.

Experimental results in Table~\ref{tab:fft_modification_table} shows accuracy drops on challenging samples that approximate half of not ensembling at all, while accuracies remain nearly the same on original samples. This supports our hypothesis that these previously inapparent high-frequency peaks revealed through diffusion noising could be captured by intermediate-step detectors to enhance detection quality on more questionable instances. Meanwhile, not applying a masked bandwidth when suppressing frequencies results in the loss of image characteristics, and decreases accuracy on original samples.

\begin{table}[t!]
    \centering
    \small
    \setlength{\tabcolsep}{3pt}
    \renewcommand{\arraystretch}{1.1}

    \begin{tabular}{ccccc}
    \toprule
\textbf{Bandwidth} & \textbf{Suppression} & \textbf{Percentile} & \textbf{Acc (\%)}     & \textbf{$\Delta$}  \\ \midrule
0.06     & 0.1  & 0.15 & 92.52/98.83 & -2.46/-0.17 \\
0.08     & 0.2  & 0.10 & 92.74/98.92 & -2.24/-0.08 \\
0.00     & 0.2  & 0.10 & 93.27/98.17 & -1.71/-0.83 \\ \bottomrule
    \end{tabular}

\caption{
    Effects of suppressing high-frequency peaks of Fourier power spectra quadrants on detection results. Models are trained on reconstructions of GenImage/GLIDE, and all the samples trained and evaluated are reconstructed through modified Fourier power spectra.
}

\label{tab:fft_modification_table}
\end{table}

\subsection{Synthetic Image Detection}\label{sec:1.2}

\begin{table}[t!]
    \centering
    \small
    \setlength{\tabcolsep}{3pt}
    \renewcommand{\arraystretch}{1.1}

    \begin{tabular}{p{2.6cm}cc}
    \toprule
\textbf{Predictions} & \textbf{Label: Synthetic} & \textbf{Label: Natural} \\ \midrule
Synthetic             & 86.95 & 9.50  \\
Natural               & 6.78  & 84.05 \\
Refuse to Answer      & 6.13  & 6.45  \\
Policy Issues         & 0.13  & 0.00  \\ \bottomrule

    \end{tabular}

\caption{
     Synthetic image detection prediction distributions of \texttt{gpt-4o} on GenImage/GLIDE original samples. Accuracies are significantly lower than current detection baselines.
}

\label{tab:gpt_detection_table}
\end{table}

\vspace{1mm}
\noindent \textbf{MLLM Detection}\quad
With the proliferation of large multimodal models, an intuitive idea would be to conduct the entire detection process with MLLMs. However, as shown in Table~\ref{tab:gpt_detection_table}, these models currently fail to achieve comparable performance with specialized detection methods, refuse to predict when facing confusing scenarios, while synthetic images may be blocked from use occasionally.

\vspace{2mm}
\noindent \textbf{Flux-generated images}\quad
Flux is an emerging generative model capable of producing high-quality images.
To test our method's performance when facing images from such generators, we generated 12,000 synthetic images with \texttt{FLUX.1-dev}, paired with natural images of the same item classes, and tested our method with the top-performing baselines on the harder and original images generated.
Our method also demonstrates state-of-the-art performance when detecting synthetic images from up-to-date generators like Flux, and leads in performance even more than on GenImage \cite{zhu2023genimage} as compared to baselines.

\begin{table}[t!]
    \centering
    \small
    \setlength{\tabcolsep}{6pt}
    \renewcommand{\arraystretch}{1.1}

    \begin{tabular}{p{1.6cm}cc}
    \toprule
\textbf{Method} & \textbf{Flux} & \textbf{GenImage Avg.} \\ \midrule
CBSID              & 73.64/99.42  & 81.64/95.95  \\
FreqNet            & 77.70/89.33  & 91.22/94.82 \\
NPR                & 92.23/98.67  & 92.43/95.85  \\
DRCT               & 84.46/98.17  & 93.19/98.36  \\
\textbf{ESIDE}     & \textbf{94.26}/\textbf{99.58}  & \textbf{95.89}/\textbf{98.91}  \\ \bottomrule

    \end{tabular}

\caption{
     Synthetic image detection accuracies on Flux.
}

\label{tab:flux_table}
\end{table}

\vspace{2mm}
\noindent \textbf{Baseline Settings}\quad
For ResNet-50 \cite{He2015DeepRL}, we initialized our model with the weights pre-trained on ImageNet, then fine-tuned it on our synthetic image detection task. For CBSID \cite{cozzolino2024raising}, we reimplemented it using the same classifier architecture as ours. For DeiT-S \cite{Touvron2020TrainingDI}, Swin-T \cite{Liu2021SwinTH}, CNNSpot \cite{Wang2019CNNGeneratedIA}, CBSID , DIRE \cite{Wang2023DIREFD}, LGrad \cite{tan2023learning}, UnivFD \cite{ojha2023towards}, FreqNet \cite{tan2024frequency}, NPR \cite{tan2024rethinking}, and DRCT \cite{chen2024drct}, we utilized the officially provided implementations from their GitHub repositories to conduct model training and evaluation.

\begin{table}[t!]
    \centering
    \small
    \setlength{\tabcolsep}{8pt}
    \renewcommand{\arraystretch}{1.1}

    \begin{tabular}{p{2.4cm}cc}
    \toprule
\textbf{Architecture} & \textbf{ESIDE}  & \textbf{CBSID}\\ \midrule
Net-XS              & 89.79/98.75       & 74.03/99.25          \\
Net-S               & 92.22/98.67           & 86.86/\textbf{99.33}         \\
Net-M               & 94.42/\textbf{99.33}  & 87.93/99.08          \\
\textbf{Net-L}      & \textbf{94.98}/99.00  & 88.53/99.17          \\
Net-XL              & 93.56/98.42           & 86.92/99.25          \\
Net-LW              & 93.34/98.50           & 86.12/\textbf{99.33}         \\ \bottomrule
    \end{tabular}

\caption{
     Comparison of effects of discriminator architectures. Models are trained on GenImage/GLIDE. 
}

\label{tab:model_architecture_table}
\end{table}

\vspace{2mm}
\noindent \textbf{Classifier Architecture}\quad
We evaluated linear network layers of varying depths and dimensions for our classifiers, to find an optimal design that balances efficiency and performance. We compared our results with the CBSID baseline of the same architecture, as shown in Table~\ref{tab:model_architecture_table}. Each network has an input size of $768$ and a scalar output, with hidden layer sizes listed below. Of all the architectures tested, Net-L performs best overall in terms of accuracy, and is used for all of our other experiments.

\vspace{1.2mm}
\begin{itemize}
    \item \textbf{Net-XS:} 256
    \item \textbf{Net-S:} 512, 256
    \item \textbf{Net-M:} 512, 256, 128
    \item \textbf{Net-L:} 1024, 512, 256, 128
    \item \textbf{Net-XL:} 1024, 1024, 512, 256, 128
    \item \textbf{Net-LW:} 2048, 1024, 512, 256
    \label{net_architectures}
\end{itemize}
\vspace{1.2mm}

\begin{table}[t!]
    \centering
    \small
    \setlength{\tabcolsep}{3pt}
    \renewcommand{\arraystretch}{1.1}

    \begin{tabular}{p{1.4cm}ccccccc}
    \toprule
\textbf{Test Split} & $s=1$ & $s=2$ & $s=3$          & $s=4$ & $s=6$ & $s=8$ & $s=12$ \\ \midrule
Hard                & 93.30 & 93.53 & \textbf{94.98} & 93.21 & 92.95 & 93.94 & 91.77  \\
Original            & 98.58 & 98.42 & \textbf{99.00} & 98.33 & 98.33 & 98.75 & 98.42  \\ \bottomrule
    \end{tabular}

\caption{
     Comparison of diffusion timestep strides on detection results. Models are trained on GenImage/GLIDE.
}

\label{tab:stride_size_table}
\end{table}

\begin{table}[t!]
    \centering
    \small
    \setlength{\tabcolsep}{1.9pt}
    \renewcommand{\arraystretch}{1.1}
\resizebox{\columnwidth}{!}{
    \begin{tabular}{p{1.2cm}ccccccccc}
    \toprule
\textbf{Timestep} & \textbf{0} & \textbf{3} & \textbf{6} & \textbf{9} & \textbf{12} & \textbf{15} & \textbf{18} & \textbf{21} & \textbf{24}\\ \midrule
Acc & \textbf{99.25} & 93.42 & 93.00 & 93.33 & 94.48 & 96.17 & 96.50 & 96.75 & 96.42 \\ \bottomrule
    \end{tabular}
}

\caption{
     Performance of 9 CBSID classification models on original samples, individually trained on noised images from different diffusion timesteps. Models are trained on GenImage/GLIDE, Net-XS architecture.
}

\label{tab:stepwise_acc_alpha_table}
\end{table}

\vspace{2mm}
\noindent \textbf{Diffusion Timestep Stride}\quad
The amount of models in the ensemble greatly affects training, to determine what timestep stride is the best for balancing performance and efficiency, we experimented with numerous ensemble sizes, as depicted in Table~\ref{tab:stride_size_table}.
Notably, optimal performance is attained when employing a stride size of $3$ instead of a unit stride. Classifiers trained on the original images typically have higher accuracy and are assigned larger model weights, as shown in Table~\ref{tab:stepwise_acc_alpha_table}. Ensembling too many noised classifiers tend to outweigh and neglect the original, thus slightly lowering accuracy. Nevertheless, when facing controversial samples, utilizing latent features enables them to coordinately overturn the unnoised discriminator’s erroneous judgment, enhancing overall accuracy.

\begin{table}[t!]
    \centering
    \small
    \setlength{\tabcolsep}{3pt}
    \renewcommand{\arraystretch}{1.2}

    \begin{tabular}{rcccc}
    \toprule
\textbf{Metric} & \textbf{Original} & \textbf{Refined-v1} & \textbf{Refined-v2} & \textbf{Refined-v3} \\ \midrule

$\overline{sim_5}$ ($\uparrow$)    &0.2192&0.2211&0.2223 &\textbf{0.2232}\\
$\overline{sim_{10}}$ ($\uparrow$) & 0.2046 & 0.2067 & 0.2081  & \textbf{0.2091}\\
$\overline{sim}$ ($\uparrow$)      & 0.1859 & 0.1871 & 0.1877  & \textbf{0.1882} \\
TTR ($\uparrow$)                   & 0.01344 & \textbf{0.01369} & 0.01368 &  \textbf{0.01369}\\
SE ($\uparrow$)                    & 0.4443 & 0.4487 & 0.4499  & \textbf{0.4509}\\
PPL ($\downarrow$)                 & \textbf{39.32} & 42.45 & 43.66  & 44.49 \\ 
\bottomrule

    \end{tabular}

\caption{
     Evaluation metrics average of each iteration during the explanation refinement process on GenExplain.
}

\label{tab:refinement_metrics_table0}
\end{table}

\subsection{Error Explanation}\label{sec:1.2}
\vspace{1mm}
\noindent \textbf{Flaw Classification} \quad
For flaw classification, a predicted label is considered ``True'' if the model's output logit is greater than or equal to $0.5$ and ``False'' otherwise. The predictions for each flaw label are computed independently.

\vspace{2mm}
\noindent \textbf{Explanation Refinement} \quad
Evaluation metric averages during explanation refinement are provided in Table~\ref{tab:refinement_metrics_table0}, while detailed values on each subset are provided in Table~\ref{tab:refinement_metrics_table1} to Table~\ref{tab:refinement_metrics_table8}.
Monotonic increases in similarity, TTR and SE indicate that refined explanations have higher relevance with the original image, and possess higher lexical diversity and information density. However, increases in PPL indicate that refining decreases fluency, attributed to the retention of specialized or domain-specific terms uncommon in general language usage.

\begin{table}[t!]
    \centering
    \small
    \setlength{\tabcolsep}{3pt}
    \renewcommand{\arraystretch}{1.2}

    \begin{tabular}{rcccc}
    \toprule
\textbf{Metric} & \textbf{Original} & \textbf{Refined-v1} & \textbf{Refined-v2} & \textbf{Refined-v3} \\ \midrule

$\overline{sim_5}$ ($\uparrow$)    &0.2206&0.2226&0.2241 &\textbf{0.2249}\\
$\overline{sim_{10}}$ ($\uparrow$) & 0.2035 & 0.2057 & 0.2073  &\textbf{ 0.2082} \\
$\overline{sim}$ ($\uparrow$)      & 0.1824 & 0.1837 & 0.1845  & \textbf{0.1850} \\
TTR ($\uparrow$)                   & 0.01882 & 0.01919 & 0.01930 &  \textbf{0.01932}\\
SE ($\uparrow$)                    & 0.4625 & 0.4679 & 0.4688  & \textbf{0.4697}\\
PPL ($\downarrow$)                 & \textbf{38.18} & 41.52 & 42.44  & 43.29 \\ \bottomrule

    \end{tabular}

\caption{
     Evaluation metrics of each iteration during the explanation refinement process on GenExplain/Midjourney.
}

\label{tab:refinement_metrics_table1}
\end{table}

\begin{table}[t!]
    \centering
    \small
    \setlength{\tabcolsep}{3pt}
    \renewcommand{\arraystretch}{1.2}

    \begin{tabular}{rcccc}
    \toprule
\textbf{Metric} & \textbf{Original} & \textbf{Refined-v1} & \textbf{Refined-v2} & \textbf{Refined-v3} \\ \midrule

$\overline{sim_5}$ ($\uparrow$)    &  0.2190 & 0.2212 & 0.2226   &  \textbf{0.2237}\\
$\overline{sim_{10}}$ ($\uparrow$) & 0.2025 & 0.2049 & 0.2063   & \textbf{0.2075} \\
$\overline{sim}$ ($\uparrow$)      & 0.1824 & 0.1837 & 0.1844   &  \textbf{0.1849}\\
TTR ($\uparrow$)                   & 0.01226 & 0.01250 &\textbf{ 0.01251}  & \textbf{0.01251}  \\
SE ($\uparrow$)                    & 0.4410 & 0.4452 & 0.4462  &\textbf{ 0.4473} \\
PPL ($\downarrow$)                 & \textbf{39.23} & 42.13 & 43.25  & 44.08 \\ \bottomrule

    \end{tabular}

\caption{
     Evaluation metrics of each iteration during the explanation refinement process on GenExplain/SD V1.4.
}

\label{tab:refinement_metrics_table3}
\end{table}

\begin{table}[t!]
    \centering
    \small
    \setlength{\tabcolsep}{3pt}
    \renewcommand{\arraystretch}{1.2}

    \begin{tabular}{rcccc}
    \toprule
\textbf{Metric} & \textbf{Original} & \textbf{Refined-v1} & \textbf{Refined-v2} & \textbf{Refined-v3} \\ \midrule

$\overline{sim_5}$ ($\uparrow$)    & 0.2217 & 0.2239 & 0.2253 & \textbf{0.2261} \\
$\overline{sim_{10}}$ ($\uparrow$) & 0.2050 & 0.2074 & 0.2089 & \textbf{0.2099} \\
$\overline{sim}$ ($\uparrow$)      & 0.1845 & 0.1858 & 0.1865 & \textbf{0.1870} \\
TTR ($\uparrow$)                   & 0.01951 & 0.01990 & 0.01992 & \textbf{0.01995} \\
SE ($\uparrow$)                    & 0.4637 & 0.4684 & 0.4701 & \textbf{0.4712} \\
PPL ($\downarrow$)                 & \textbf{40.32} & 43.55 & 44.84 & 45.58 \\ \bottomrule

    \end{tabular}

\caption{
     Evaluation metrics of each iteration during the explanation refinement process on GenExplain/SD V1.5.
}

\label{tab:refinement_metrics_table2}
\end{table}

\begin{table}[t!]
    \centering
    \small
    \setlength{\tabcolsep}{3pt}
    \renewcommand{\arraystretch}{1.2}

    \begin{tabular}{rcccc}
    \toprule
\textbf{Metric} & \textbf{Original} & \textbf{Refined-v1} & \textbf{Refined-v2} & \textbf{Refined-v3} \\ \midrule

$\overline{sim_5}$ ($\uparrow$)    & 0.2181 & 0.2197 & 0.2207 & \textbf{0.2213}  \\
$\overline{sim_{10}}$ ($\uparrow$) & 0.2057 & 0.2075 & 0.2087 & \textbf{0.2094} \\
$\overline{sim}$ ($\uparrow$)      &  0.1890 & 0.1899& 0.1904& \textbf{0.1908}  \\
TTR ($\uparrow$)                   & 0.01272 &\textbf{0.01280} &0.01277 &0.01273\\
SE ($\uparrow$)                    & 0.4420 & 0.4458& 0.4468& \textbf{0.4479} \\
PPL ($\downarrow$)                 & \textbf{39.78} & 42.67& 43.74& 44.81 \\ \bottomrule

    \end{tabular}

\caption{
     Evaluation metrics of each iteration during the explanation refinement process on GenExplain/ADM.
}

\label{tab:refinement_metrics_table5}
\end{table}

\begin{table}[t!]
    \centering
    \small
    \setlength{\tabcolsep}{3pt}
    \renewcommand{\arraystretch}{1.2}

    \begin{tabular}{rcccc}
    \toprule
\textbf{Metric} & \textbf{Original} & \textbf{Refined-v1} & \textbf{Refined-v2} & \textbf{Refined-v3} \\ \midrule

$\overline{sim_5}$ ($\uparrow$)    &0.2233 &0.2253& 0.2265& \textbf{0.2275}\\
$\overline{sim_{10}}$ ($\uparrow$) &0.2087& 0.2109& 0.2123& \textbf{0.2133}\\
$\overline{sim}$ ($\uparrow$)      &0.1900& 0.1912& 0.1918&\textbf{ 0.1923}\\
TTR ($\uparrow$)                   &  0.01064 &\textbf{ 0.01085}& 0.01074& 0.01072  \\
SE ($\uparrow$)                    &  0.4334 &0.4380& 0.4390& \textbf{0.4400}   \\
PPL ($\downarrow$)                 & \textbf{38.50}& 41.31 &42.40 &43.12  \\ \bottomrule

    \end{tabular}

\caption{
     Evaluation metrics of each iteration during the explanation refinement process on GenExplain/GLIDE.
}

\label{tab:refinement_metrics_table4}
\end{table}

\begin{table}[t!]
    \centering
    \small
    \setlength{\tabcolsep}{3pt}
    \renewcommand{\arraystretch}{1.2}

    \begin{tabular}{rcccc}
    \toprule
\textbf{Metric} & \textbf{Original} & \textbf{Refined-v1} & \textbf{Refined-v2} & \textbf{Refined-v3} \\ \midrule

$\overline{sim_5}$ ($\uparrow$)    & 0.2203& 0.2226& 0.2239  & \textbf{0.2248} \\
$\overline{sim_{10}}$ ($\uparrow$) & 0.2052&0.2075& 0.2090 & \textbf{0.2100} \\
$\overline{sim}$ ($\uparrow$)      &  0.1861 &0.1873 &0.1879 & \textbf{0.1885} \\
TTR ($\uparrow$)                   & 0.01227 &0.01239& 0.01239  &\textbf{0.01245} \\
SE ($\uparrow$)                    & 0.4420& 0.4453& 0.4465 & \textbf{0.4477} \\
PPL ($\downarrow$)                 &  \textbf{40.11}& 43.05& 44.26   &  45.08\\ \bottomrule

    \end{tabular}

\caption{
     Evaluation metrics of each iteration during the explanation refinement process on GenExplain/Wukong.
}

\label{tab:refinement_metrics_table6}
\end{table}

\begin{table}[t!]
    \centering
    \small
    \setlength{\tabcolsep}{3pt}
    \renewcommand{\arraystretch}{1.2}

    \begin{tabular}{rcccc}
    \toprule
\textbf{Metric} & \textbf{Original} & \textbf{Refined-v1} & \textbf{Refined-v2} & \textbf{Refined-v3} \\ \midrule

$\overline{sim_5}$ ($\uparrow$)    &  0.2161& 0.2177& 0.2187 &\textbf{0.2195} \\
$\overline{sim_{10}}$ ($\uparrow$) & 0.2036 &0.2053& 0.2065& \textbf{0.2074}\\
$\overline{sim}$ ($\uparrow$)      &  0.1856& 0.1867 &0.1872 &\textbf{0.1876}   \\
TTR ($\uparrow$)                   & 0.01050& \textbf{0.01077}& 0.01075& 0.01076  \\
SE ($\uparrow$)                    &  0.4341& 0.4389& 0.4401&\textbf{0.4410} \\
PPL ($\downarrow$)                 & \textbf{39.40} &42.92& 44.37& 45.17  \\ \bottomrule

    \end{tabular}

\caption{
     Evaluation metrics of each iteration during the explanation refinement process on GenExplain/VQDM.
}

\label{tab:refinement_metrics_table7}
\end{table}

\begin{table}[t!]
    \centering
    \small
    \setlength{\tabcolsep}{3pt}
    \renewcommand{\arraystretch}{1.2}

    \begin{tabular}{rcccc}
    \toprule
\textbf{Metric} & \textbf{Original} & \textbf{Refined-v1} & \textbf{Refined-v2} & \textbf{Refined-v3} \\ \midrule

$\overline{sim_5}$ ($\uparrow$)    &  0.2142 &0.2158& 0.2169 &\textbf{0.2177}   \\
$\overline{sim_{10}}$ ($\uparrow$) &  0.2028 &0.2047 &0.2059 &\textbf{0.2068} \\
$\overline{sim}$ ($\uparrow$)      &  0.1870 &0.1881& 0.1888&\textbf{0.1892}   \\
TTR ($\uparrow$)                   &  0.01083& \textbf{0.01109} &0.01105 &0.01106  \\
SE ($\uparrow$)                    & 0.4360 &0.4402 &0.4413& \textbf{0.4420}  \\
PPL ($\downarrow$)                 & \textbf{39.04}& 42.47& 43.98 &44.80  \\ \bottomrule

    \end{tabular}

\caption{
     Evaluation metrics of each iteration during the explanation refinement process on GenExplain/BigGAN.
}

\label{tab:refinement_metrics_table8}
\end{table}

\onecolumn
\newpage
\section{Prompt Design}\label{sec:b}

\subsection{Flaw Classification Prompt}\label{sec:b.1}

\vspace{2mm}
To generate the initial set of classified image flaws for constructing our \scalebox{1.1}{G}\scalebox{0.8}{EN}\scalebox{1.1}{E}\scalebox{0.8}{XPLAIN} dataset, and then train a flaw classification model, we first prompted \texttt{GPT-4o} to classify these flaws, which were manually filtered afterwards.

\vspace{2mm}
\begin{enumerate}[label=,left=-0.4cm,ref=\theenumi]
\item 
\centering
\begin{tikzpicture}
    \node[fill=orange!60, draw, font=\bfseries, text width=0.95\textwidth, align=justify, inner sep=10pt,text=white, rounded corners] at (0, 0)     
    {Flaw Classification Guidelines for Synthetic Images};
    
    \node[fill=orange!5, draw, text width=0.95\textwidth, align=justify, inner sep=10pt, rounded corners] at (0, -8.86) 
    { \setstretch{1.2}
    You will be provided with an AI generated image, try to identify systematic errors that make it distinguishable from natural real images from the categories below: \\
    
    \textbf{1. Lighting:} Unnatural or inconsistent light sources and shadows.
    
    \textbf{2. Color Saturation or Contrast:} Overly bright or dull colors, extreme contrasts disrupting image harmony.
    
    \textbf{3. Perspective:} Spatial disorientation caused by unrealistic angles or viewpoints, otherwise dimensionality errors such as flattened 3D objects.
    
    \textbf{4. Bad Anatomy:} For living creatures, mismatches and errors of body parts in humans or animals.
    
    \textbf{5. Distorted Objects:} For nonliving objects only, warped objects with fallacious details deviating from expected forms.
    
    \textbf{6. Structural Composition:} Poor positional arrangement between multiple elements in the scene.

    \textbf{7. Incomprehensible Text:} Malformed and unrecognizable text.
    
    \textbf{8. Implausible Scenarios:} Inappropriate behavior and situations unlikely to happen based on sociocultural concerns, or contradicting to historical facts.
    
    \textbf{9. Physical Law Violations:} Improbable physics, such as erroneous reflections or objects defying gravity. \\
    
    Only select the categories below when highly confident: \\

    \textbf{10. Blurry or Inconsistent Borders:} Unclear or abrupt border outlines between elements.
    
    \textbf{11. Background:} Poorly blended or monotonous drab backgrounds.
    
    \textbf{12. Texture:} For nonliving objects only, significantly over-polished or unnatural textural appearances.
    
    \textbf{13. Generation Failures:} Major prominent rendering glitches or incomplete objects disrupting the entire scene.\\

    If NONE of the categories above match, select: \\

    \textbf{14. Not Evident} \\

    Choose one or more categories above, and only reply the indexes of identified errors separated with commas, e.g. "1,3,6" for an image with "Lighting, Perspective, Structural Composition" errors, with no additional explanation. \\
    
    PLEASE NOTE: Responses MUST be in the format of "[Number 1],[Number 2],..." (numbers and commas only) ordered from low to high, or "14" if no evident errors are found.

    };

    \node[fill=orange!30, draw=none, font=\bfseries, text width=0.95\textwidth+16.4pt, align=justify, inner sep=1.6pt,] at (0,-0.34){};
\end{tikzpicture}
\end{enumerate}

\newpage

\subsection{Initial Explanation Generation Prompt}\label{sec:b.2}

\vspace{2mm}
After manually removing images with flaws that were incorrectly categorized, we prompted \texttt{GPT-4o-mini} to generate an initial explanation based on the description of the identified flaw.

\vspace{2mm}
\begin{enumerate}[label=,left=-0.4cm,ref=\theenumi]
\item 
\centering
\begin{tikzpicture}
    \node[fill=orange!60, draw, font=\bfseries, text width=0.95\textwidth, align=justify, inner sep=10pt,text=white, rounded corners] at (0, 0)     
    {Initial Explanation Generation Guidelines};
    
    \node[fill=orange!5, draw, text width=0.95\textwidth, align=justify, inner sep=10pt, rounded corners] at (0, -2.49) 
    { \setstretch{1.2}
        You will be provided with an AI generated image confirmed with the following systematic error: $<$DESCRIPTION OF ERROR CATEGORY$>$ \\

        Give an explanation for why such an error is found in the image, and point out the specific location or items causing the error. Pay close attention to image details. \\
        
        PLEASE NOTE: Responses should be concise, and organized into a SINGLE PARAGRAPH.
    };

    \node[fill=orange!30, draw=none, font=\bfseries, text width=0.95\textwidth+16.4pt, align=justify, inner sep=1.6pt,] at (0,-0.34){};
\end{tikzpicture}
\end{enumerate}

\vspace{3mm}
\subsection{Iterative Explanation Refinement Prompt}\label{sec:b.3}

\vspace{2mm}
For 3 rounds we iteratively refined the initial explanation and follow-ups. During each iteration, we calculated the Top-10 similar noun phrases with the synthetic image, and prompted \texttt{GPT-4o-mini} to retain these relevant phrases while additionally searching for overlooked flaws. The final refined explanations were added to our dataset.

\vspace{2mm}
\begin{enumerate}[label=,left=-0.4cm,ref=\theenumi]
\item 
\centering
\begin{tikzpicture}
    \node[fill=orange!60, draw, font=\bfseries, text width=0.95\textwidth, align=justify, inner sep=10pt,text=white, rounded corners] at (0, 0)     
    {Iterative Explanation Refinement Guidelines};
    
    \node[fill=orange!5, draw, text width=0.95\textwidth, align=justify, inner sep=10pt, rounded corners] at (0, -3.51) 
    { \setstretch{1.2}
        You will be provided with an AI generated image confirmed with $<$ERROR CATEGORY$>$ error. Below is an explanation of why this error appears in the image: $<$PREVIOUS EXPLANATION$>$ \\
        
        In this explanation, the following words may be highly relevant and accurately describe the image: $<$TOP 10 SIMILARITY PHRASES$>$ \\
        
        Your task is to refine the explanation to better align with the error in the image while retaining the relevant words. You can also analyze the image to identify whether any other potential errors that may have been overlooked about $<$ERROR CATEGORY$>$. Keep the explanation concise and avoid redundancy. \\
        
        PLEASE NOTE: Responses should be concise, and organized into a SINGLE PARAGRAPH.
    };

    \node[fill=orange!30, draw=none, font=\bfseries, text width=0.95\textwidth+16.4pt, align=justify, inner sep=1.6pt,] at (0,-0.34){};
\end{tikzpicture}
\end{enumerate}

\end{document}